\documentclass[10pt,twocolumn,letterpaper]{article}

\usepackage[pagenumbers]{iccv} 

\usepackage{times}
\usepackage{epsfig}
\usepackage{graphicx}
\usepackage{amsmath}
\usepackage{amssymb}
\usepackage{booktabs, makecell, tabularx}
\usepackage{multirow}
\usepackage{dsfont}
\usepackage{enumitem}
\usepackage{algorithm}
\usepackage{algpseudocode}
\usepackage{xspace}

\usepackage{animate}
\usepackage{booktabs} 

\usepackage{booktabs}
\usepackage[normalem]{ulem}
\usepackage{ifthen}
\usepackage{epsfig}
\usepackage{graphics}
\usepackage{graphicx}
\usepackage{amsmath}
\usepackage{animate}
\usepackage{amsmath}
\usepackage{booktabs}
\usepackage{hhline}
\usepackage{adjustbox}
\usepackage{multirow}
\usepackage{booktabs,arydshln}
\usepackage{amsfonts}

\usepackage{placeins}
\usepackage{amsmath}
\usepackage{amssymb}
\usepackage{booktabs}
\usepackage{makecell}
\usepackage{colortbl}
\usepackage{url}
\usepackage{float}
\usepackage{subcaption}
\usepackage{graphicx}
\usepackage{wrapfig}
\usepackage{xcolor} 
\usepackage{framed}
\usepackage{color}
\usepackage{enumitem}
\usepackage{multirow}
\usepackage{arydshln} 
\usepackage{booktabs}
\usepackage{bbding}
\usepackage{pifont}
\makeatletter
\@namedef{ver@everyshi.sty}{}
\makeatother
\usepackage{tikz}
\definecolor{mediumpurple}{RGB}{30,144,255}
\definecolor{thistle}{RGB}{216, 191, 216}
\definecolor{lightgreen}{RGB}{188,238,104}
\definecolor{lightblue}{RGB}{191,239,255}
\definecolor{lightcoral}{RGB}{240,128,128}
\definecolor{pink}{RGB}{255,192,203}
\newcommand{\mixshape}{\raisebox{0.5pt}{\tikz\fill[thistle] (0,0) circle (.8ex);}}
\newcommand{\proshape}{\raisebox{0.5pt}{\tikz\fill[lightcoral] (0,0) circle (.8ex);}}
\newcommand{\fineshape}{\raisebox{0.5pt}{\tikz\fill[lightblue] (0,0) circle (.8ex);}}
\newcommand{\ensshape}{\raisebox{0.5pt}{\tikz\fill[pink](0,0) circle (.8ex);}}
\newcommand{\stshape}{\raisebox{0.5pt}{\tikz\fill[lightgreen](0,0) circle (.8ex);}}

\definecolor{iccvblue}{rgb}{0.21,0.49,0.74}
\usepackage[pagebackref,breaklinks,colorlinks,allcolors=iccvblue]{hyperref}

\makeatletter
\newcommand{\ostar}{\mathbin{\mathpalette\make@circled\star}}
\newcommand{\make@circled}[2]{%
  \ooalign{$\m@th#1\smallbigcirc{#1}$\cr\hidewidth$\m@th#1#2$\hidewidth\cr}%
}
\newcommand{\smallbigcirc}[1]{%
  \vcenter{\hbox{\scalebox{0.77778}{$\m@th#1\bigcirc$}}}%
}

\def\paperID{6755} 
\def\confName{ICCV}
\def\confYear{2025}

\title{REGEN: Learning Compact Video Embedding with (Re-)Generative Decoder}

\author{
  Yitian Zhang$^{1,2}$\textsuperscript{\dag}\ \ \
  Long Mai$^{1}$\ \ \ 
  Aniruddha Mahapatra$^{1}$\ \ \ 
  David Bourgin$^{1}$\\
  Yicong Hong$^{1}$\ \ \ 
  Jonah Casebeer$^{1}$\ \ \ 
  Feng Liu$^{1}$\ \ \ 
  Yun Fu$^{2}$\\[1.5ex]
    $^{1}$Adobe Research\ \ \ 
    $^{2}$Northeastern University\\[2ex]
\url{https://bespontaneous.github.io/REGEN/}
}

\begin{document}
\twocolumn[{%
\maketitle
\renewcommand\twocolumn[1][]{#1}%

\vskip -0.2in
    \centering
    \begin{minipage}{\textwidth}
        \includegraphics[width=\linewidth]{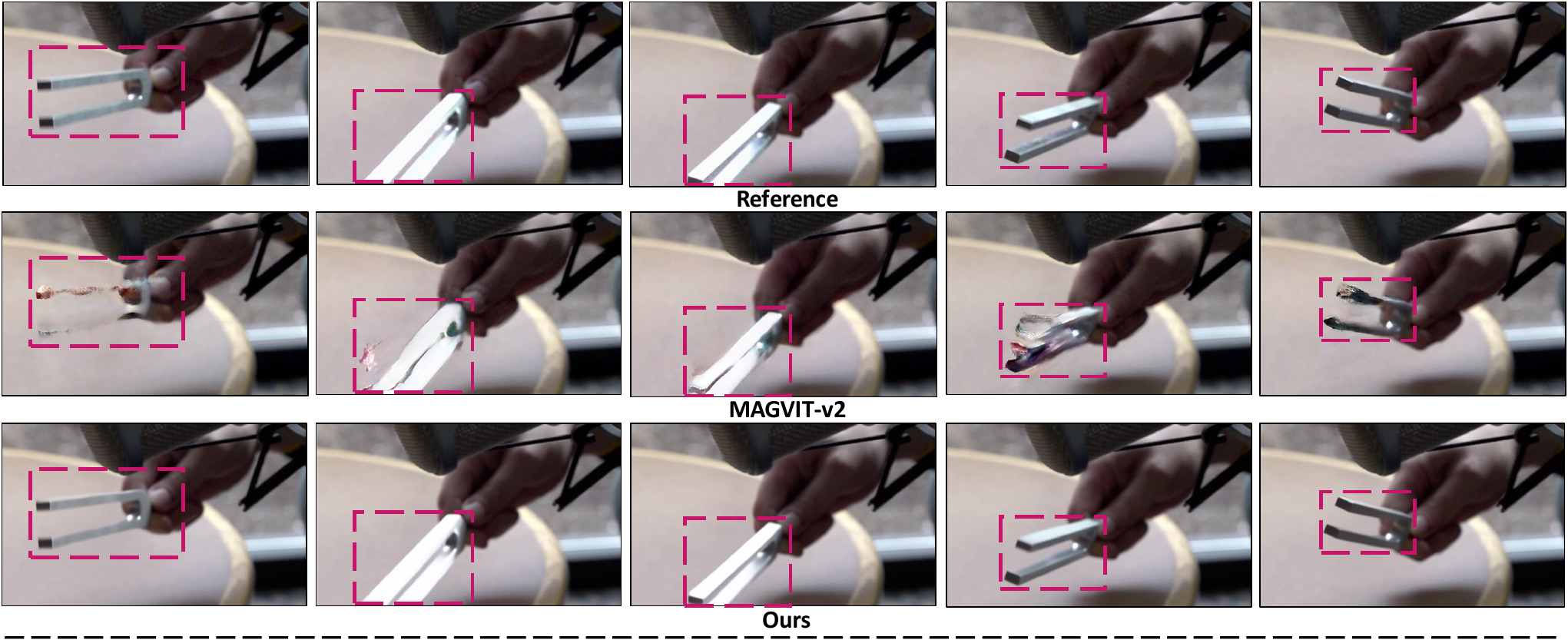}
    \end{minipage}
    \vskip 0.02in
    \begin{minipage}{0.325\textwidth}
        \animategraphics[autoplay,loop,width=\linewidth]{24}{Graphics/t2v_6_long/}{1}{66}
        \vskip -0.02in
        \par\small\centering \textit{Prompt: A sheep behind a fence looking at the camera.}
    \end{minipage}%
    \hfill
    \begin{minipage}{0.325\textwidth}
        \animategraphics[autoplay,loop,width=\linewidth]{24}{Graphics/t2v_2_long/}{1}{66}
        \vskip -0.02in
        \par\small\centering \textit{Prompt: Young woman with red hair and leafy crown.}
    \end{minipage}%
    \hfill
    \begin{minipage}{0.325\textwidth}
        \animategraphics[autoplay,loop,width=\linewidth]{24}{Graphics/t2v_4_long/}{1}{66}
        \vskip -0.02in
        \par\small\centering \textit{Prompt: Time lapse at the snow land with aurora in the sky.}
    \end{minipage}
\vskip -0.1in
\captionof{figure}{Reconstruction (top) and
text-to-video (T2V) generations (bottom) results at 32$\times$ temporal compression.
This figure contains video results (for T2V results), best viewed with Adobe Acrobat Reader.
Colored bounding boxes in reconstruction results denote the regions with the most difference where MAGVIT-v2 leads to clear artifacts in the tuning fork.
}
\label{fig:teaser}
\vskip 0.1in
}]


\renewcommand{\thefootnote}{\dag}
\footnotetext{This work was done during an internship at Adobe Research.}
\renewcommand{\thefootnote}{\arabic{footnote}}

\begin{abstract}
We present a novel perspective on learning video embedders for generative modeling: rather than requiring an exact reproduction of an input video, an effective embedder should focus on synthesizing visually plausible reconstructions. This relaxed criterion enables substantial improvements in compression ratios without compromising the quality of downstream generative models. Specifically, we propose replacing the conventional encoder-decoder video embedder with an encoder-generator framework that employs a diffusion transformer (DiT) to synthesize missing details from a compact latent space. 
Therein, we develop a dedicated latent conditioning module to condition the DiT decoder on the encoded video latent embedding. Our experiments demonstrate that our approach enables superior encoding-decoding performance compared to state-of-the-art methods, particularly as the compression ratio increases. To demonstrate the efficacy of our approach, we report results from our video embedders achieving a temporal compression ratio of up to 32× (8× higher than leading video embedders) and validate the robustness of this ultra-compact latent space for text-to-video generation, providing a significant efficiency boost in latent diffusion model training and inference.
\end{abstract}

\vspace{-0.2in}
\section{Introduction}
\label{sec:intro}

Diffusion models~\cite{sohl2015deep,ho2020denoising} have recently emerged as a major paradigm in generative content creation, achieving breakthroughs across various domains such as images~\cite{ho2022imagen,ho2022video,peebles2023scalable,flux,rombach2022high}, videos~\cite{sora,runwaygen3,kling,opensora,yang2024cogvideox,menapace2024snap}, and audio~\cite{kreuk2022audiogen,copet2024simple,huang2302noise2music,liu2023audioldm}. 
For efficiency, most modern diffusion models compress the input media into a compact latent space before modeling, commonly referred to as latent diffusion models (LDMs)~\cite{rombach2022high}.
A core component of the LDM is the embedding model (which we refer to as \textbf{embedder} in this paper), which typically follows an encoder-decoder architecture. The encoder transforms the raw data into a compact latent representation, while the diffusion models operate within this latent space and subsequently transform the data back to the raw space using the decoder.

For video generation, state-of-the-art (SOTA) video embedder, exemplified by the continuous-token variants of the MAGVIT-v2 model~\cite{yu2023language}, leverage architectures that enable spatiotemporal processing, offering compression in both spatial and temporal dimensions. Despite these advances, SOTA video embedders typically offer 8$\times$ spatial compression but only 4$\times$ temporal compression.

Increasing compression is inherently challenging due to a fundamental trade-off between representational efficiency and reconstruction fidelity. At high compression ratios, the model struggles to retain sufficient information within the latent space to reproduce high-fidelity details accurately. 
We verify that adapting existing video embedders to support temporal compression beyond 4$\times$ is highly challenging without significantly degrading reconstruction quality.

This paper presents an alternative perspective on learning video tokenizers tailored for latent diffusion modeling. We propose a generation-oriented view of latent representation learning.
``\textit{We argue that, in the context of latent diffusion, the key desired property of the latent space is the ability to generate visually plausible content rather than to faithfully recover an input video.}''
We propose transforming the traditional encoder-decoder video embedder into an encoder-generator. This generation-oriented perspective enables a more flexible approach to allowing high-compression in the latent space. 
By leveraging generative capabilities in the decoder, the encoder can focus only on preserving the essential semantic and structural information. The decoder then synthesizes realistic finer details.

To this end, we introduce \textbf{REGEN}, an approach for learning a high-compression video embedder via a diffusion decoder with a resynthesis objective.
REGEN utilizes a \textbf{diffusion Transformer (DiT)} \cite{peebles2023scalable} as the video decoder due to its exceptional video modeling capability compared to diffusion U-Net, as consistently demonstrated by impactful works \cite{sora,opensora,pku_yuan_lab_and_tuzhan_ai_etc_2024_10948109,menapace2024snap}. Our DiT decoder treats the latent features encoded from an input video sequence as the conditioning signal during the diffusion-based generation process. Both the encoder and the DiT decoder are jointly trained within the diffusion-based training framework to learn to generate the original video content.

To connect the encoded latent features with the DiT decoder, we design a dedicated \textbf{latent conditioning module}. This module introduces a novel DiT conditioning mechanism that transforms the latent maps into a content-aware positional embedding, enabling strong conditioning signals for the decoder's diffusion process. 
Our proposed latent conditioning mechanism also naturally enables continuous-time decoding that supports not only reconstruction but also interpolation and extrapolation capabilities. More importantly, it also addresses the challenge of encoding and decoding for arbitrary resolutions and aspect ratios, unseen during training, which is difficult to achieve with DiT.

In this work, we aim to enhance the compactness of the latent space beyond the current limits and
focus on the axis of increasing the temporal compression of our embedder, keeping the spatial compression fixed at 8$\times$. 
Across all the experiments, we follow the design of MAGVIT-v2 \cite{yu2023language} to keep the number of latent channels to be 8 for all compression ratios,
as we want to isolate the effect of increasing temporal compression on reconstruction quality.

\begin{figure*}[t]
\begin{center}
\scalebox{1.0}{\includegraphics[width=\textwidth]{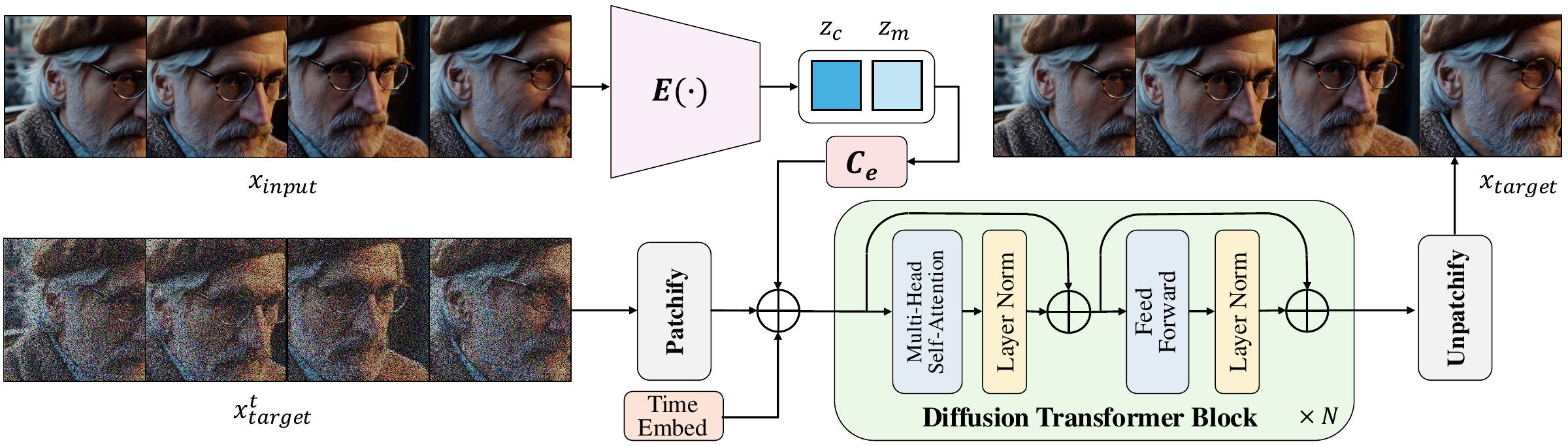}}
\end{center}
\vskip -0.2in
\caption{
\textbf{Overall framework.}
Our spatiotemporal video encoder $E\left ( \cdot \right )$ encodes the input video $x_{input}$ into two latent frames, content and motion $\left ( z_{c}, z_{m} \right )$. They are processed by the latent expansion module $C_{e}$ and serve as conditioning for the generative decoder.
}
\label{fig:method}
\end{figure*}

The contributions of the current work are as follows: 
\begin{enumerate}
    \item We present the idea of using a Diffusion Transformer as a decoder in video modeling instead of the conventional VAE~\cite{kingma2013auto}-based encoder-decoder. We demonstrate that this modeling strategy allows a re-design of the learning objective in video encoding, allowing us to bypass the fundamental compression-reconstruction trade-off.  
    \item We introduce a novel latent conditioning module with a content-aware positional encoding formulation. This enables us to effectively turn the encoded latent features into spatiotemporal control signals for the Diffusion Transformer (DiT) decoder. 
    Our decoder not only enables DiT to encode and decode videos of arbitrary aspect ratios and resolutions, but also supports even one-step sampling without utilizing external distillation.
    \item We show that REGEN enables superior reconstruction performance compared to extending the VAE-based embedders at high compression ratios. Our method even outperforms even very recent SOTA video embedders at commonly used 4$\times$ temporal compression that have the same number of latent channels as our method. We also show that our highest-compression (32$\times$ temporal) latent space is friendly for text-to-video generation.
\end{enumerate}

\vspace{-0.05in}
\section{Related work}
\label{sec:relw}

\noindent \textbf{Video Embedders.} Early video diffusion models adapted image LDM, using the same image-based embedder for per-frame latent feature extraction \cite{he2022latent,singer2022make,luo2023videofusion,ge2023preserve,an2023latent,wang2023videofactory}. 
This frame-wise approach neglects temporal relationships, causing inconsistencies and restricting the compactness of the video latent.
Recent video diffusion frameworks utilize spatiotemporal embedders for joint processing. The seminal work of MAGVIT-v2~\cite{yu2023language} employs causal 3D convolutions for compression across both spatial and temporal dimensions. 
Originally designed for discrete tokenization, MAGVIT-v2 has been adapted for continuous-token video encoding, facilitating its integration into various prominent video diffusion models \cite{yang2024cogvideox,opensora,pku_yuan_lab_and_tuzhan_ai_etc_2024_10948109}.
Despite their effectiveness, they typically offer limited temporal compression (e.g., 4$\times$). 
We aim to enhance the compactness of the latent space beyond the current limits, extending up to 32$\times$ temporal compression. 
To accomplish it, we offer a novel perspective of using encoder-generator with a (re-)generative diffusion decoding process to escape the reconstruction-compression trade-off.

\noindent \textbf{Diffusion Autoencoders.} Our work draws inspiration from recent studies that demonstrate the potential of image diffusion models to generate content conditioned on abstract feature vectors extracted from conditioning images \cite{betker2023improving,preechakul2022diffusion,hudson2024soda}. Specifically, \cite{preechakul2022diffusion} shows that an image diffusion model can be semantically guided, akin to style-code manipulation in StyleGAN \cite{karras2019style}. Similarly, \cite{betker2023improving} validates the improvement brought by generative decoders, and \cite{hudson2024soda} highlights the ability to reconstruct images from conditioning feature vectors. 
We adopt this latent-conditioning diffusion concept to the context of video embedder learning.
Complementary to them, to the best of our knowledge, we are the first to explore diffusion transformer autoencoders in the context of learning highly compact latent spaces for videos.

\vspace{-0.05in}
\section{Method}
\label{sec:method}

Our model consists of two major components.
First, the spatiotemporal video encoder that projects the input video sequence into a compact latent space (Section \ref{sec:encoder}).
The second is the DiT-based generative decoder with that converts the latents back to pixels by taking the latents as a conditioning signal (Section \ref{sec:decoder}). The whole model is trained from scratch in an end-to-end fashion with the diffusion objective. Figure ~\ref{fig:method} illustrates our overall framework.

\subsection{Spatiotemporal Video Encoder}
\label{sec:encoder}

The goal of our encoder is to encode a video into a compact latent space. 
Inspired by MAGVIT-v2~\cite{yu2023language}, we adopt the continuous version of their encoder design due to its ability to encode both images and videos in consistent latent space. 

The encoder consists of multiple 3D convolution blocks, which are causal in the temporal dimension.
For a video of $T+1$ frames with dimensions $H \times W$, at a spatial downsampling factor of $m$, and a temporal downsampling rate of $k$, due to the causal nature, the encoder produces a sequence of $1+\frac{T}{k}$ latent feature maps each of dimension $\frac{H}{m}\times \frac{W}{m}$.
To provide flexibility in consistently encoding long videos without going out-of-memory, we follow a chunk-wise encoding scheme that encodes videos of fixed-length frames (a \textit{chunk}). 
Thus, at a desired temporal compression ratio $k$, we encode any input video chunk $x_{input}$ of length $k+1$ into two latent frames:
\begin{equation}
    z_{c}, z_{m} = E\left ( x_{input} \right ),
\end{equation}
where $z_{c}$ and $z_{m}$ denote the resulting two latent maps. 
As the encoder is causal in nature, $z_c$ only contains information from the first frame. We call this the \textbf{content} latent frame. 
$z_m$, contains the compressed motion information of the rest of the frames. We refer to this as the \textbf{motion} latent frame.
We use 8 latent channels for both $z_{c}$ and $z_{m}$.

\begin{figure}[t]
\begin{center}
\scalebox{0.48}{\includegraphics[width=\textwidth]{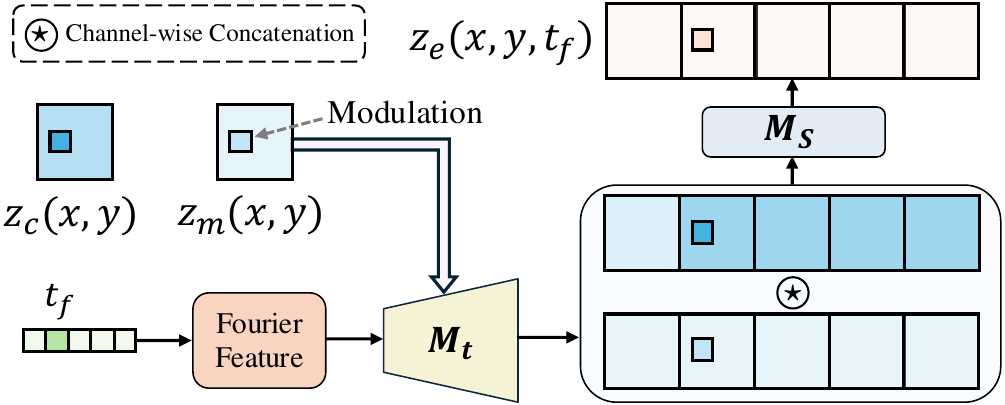}}
\end{center}
\vskip -0.2in
\caption{
\textbf{Latent conditioning module $C_{e}$.} 
The SIREN network $M_{t}$ maps the time coordinate $t_{f}$ to a feature vector modulated by the motion latent $z_{m}$. The resulting feature is concatenated with the feature value of $z_c$ at the corresponding spatial coordinate $(x,y)$. The concatenated feature is mapped into the DiT hidden dimension by the projector $M_{s}$. We utilize the first frame prediction from SIREN to replace the first frame of expanded $z_{c}$ to ensure consistent representation for both image and video inputs.
}
\label{fig:condition}
\end{figure}

\subsection{Diffusion Transformer Decoder}
\label{sec:decoder}

As opposed to the conventional convolution-based decoder, we use a transformer~\cite{vaswani2017attention} and model the decoding task as a conditional diffusion process.
Given the input video sequence $x_{input}$ and its corresponding content and motion latents $[z_{c}, z_{m}]$, the diffusion decoder $G_{d}$ is trained to (re-)generate a target video output $x_{target}$ from a noise map sequence $N$, conditioned on the latent maps $[z_{c}, z_{m}]$. For the reconstruction task, $x_{target}$ is chosen to be the same as $x_{input}$. It's worth noting that this formulation inherently allows for a more flexible definition of $x_{target}$ to handle different tasks beyond pure reconstruction. For example, setting $x_{target}$ to be the temporally upsampled version of $x_{input}$ corresponds to modeling the temporal interpolation, while setting $x_{target}$ to be a time-shifted version of $x_{input}$ leads to the temporal extrapolation.

During training, we corrupt the clean data $x_{target}$ (denoting as $x_{target}^{0}$ for diffusion noise timestep $t=0$) with the forward diffusion process: 
\begin{equation}
  q(x_{target}^t | x_{target}^0) = \mathcal{N}(x_{target}^t; \sqrt{\alpha^t} x_{target}^0, (1 - \alpha^t) I),
\end{equation}
where $t$ is the diffusion time step, and $\alpha^t$ is determined via the noise scheduling procedure \cite{liu2023instaflow}.
The sample at time step $t$ is obtained with the reparameterization trick:
\begin{equation}
  x_{target}^t = \sqrt{\alpha^t} x_{target}^0 + \sqrt{1 - \alpha^t} \, \epsilon,
\end{equation}
where $\epsilon \sim \mathcal{N}(0, I)$.
In the reverse denoising process, the generative decoder learns to invert the forward corruption under the condition $[z_c, z_m]$. Specifically, a denoising model $\epsilon_\theta$ with parameters $\theta$ is trained to predict the noise $\epsilon$ at each time step $t$ given the corresponding noisy version $x_{target}^{t}$, conditioning on $[z_c, z_m]$ with the simplified learning objective:
\begin{equation}
  \mathcal{L}(\theta) = \left\| \epsilon - \epsilon_\theta(x^t_{target}, [z_c, z_m]) \right\|^2,
\end{equation}
Under this formulation, we train our video encoder along with the generative decoder in an end-to-end fashion.
Our diffusion decoder leverages a DiT architecture for the denoiser backbone due to its superior modeling and scaling capabilities compared to U-Net as pointed to many prior works~\cite{sora,opensora,pku_yuan_lab_and_tuzhan_ai_etc_2024_10948109,menapace2024snap}.
Our decoder operates in pixel space, at a patch size of $p$. The patch size $p$ is determined according to the spatial down-sampling ratio in the encoder, which in our experiments is 8$\times$. Thus, we keep $p=8$.

\begin{table*}[h]
\centering
\scalebox{0.8}{\begin{tabular}{lccccccccccccc}
\toprule
\multirow{3}*{Method} & \multirow{3}*{Compression} & \multicolumn{6}{c}{MCL-JCV} & \multicolumn{6}{c}{DAVIS 2019} \\
\cmidrule(lr){3-8}\cmidrule(lr){9-14}
& & \multicolumn{3}{c}{256$\times$256} & \multicolumn{3}{c}{512$\times$512} & \multicolumn{3}{c}{256$\times$256} & \multicolumn{3}{c}{512$\times$512} \\
\cmidrule(lr){3-5}\cmidrule(lr){6-8}\cmidrule(lr){9-11}\cmidrule(lr){12-14}
& & PSNR & SSIM & rFVD$\downarrow$ & PSNR & SSIM & rFVD$\downarrow$ & PSNR & SSIM & rFVD$\downarrow$ & PSNR & SSIM & rFVD$\downarrow$ \\
\midrule
\stshape{} MAGVIT-v2  & \multirow{2}*{8$\times$8$\times$8} & 26.61 & 0.718 & 105.72 & 29.14 & 0.771 & 72.07 & 22.82 & 0.602 & 183.52 & 24.75 & 0.660 & 125.03 \\
\mixshape{} REGEN  & & \textbf{28.82} & \textbf{0.785} & \textbf{85.37} & \textbf{32.74} & \textbf{0.846} & \textbf{29.88} & \textbf{26.00} & \textbf{0.711} & \textbf{152.46} & \textbf{29.34} & \textbf{0.778} & \textbf{89.98} \\
\midrule
\stshape{} MAGVIT-v2  & \multirow{2}*{8$\times$8$\times$16} & 25.06 & 0.672 & 205.75 & 26.62 & 0.717 & 185.69 & 20.62 & 0.527 & 441.24 & 21.21 & 0.572 & 417.43 \\
\mixshape{} REGEN  & & \textbf{27.27} & \textbf{0.736} & \textbf{174.29} & \textbf{30.41} & \textbf{0.798} & \textbf{92.48} & \textbf{23.85} & \textbf{0.635} & \textbf{328.83} & \textbf{26.27} & \textbf{0.699} & \textbf{235.13} \\
\midrule
\stshape{} MAGVIT-v2  & \multirow{2}*{8$\times$8$\times$32} & 22.97 & 0.573 & 536.01 & $^{\dagger}$ & $^{\dagger}$ & $^{\dagger}$ & 18.23 & 0.419 & 1080.15 & $^{\dagger}$ & $^{\dagger}$ & $^{\dagger}$ \\
\mixshape{} REGEN  & & \textbf{26.05} & \textbf{0.695} & \textbf{265.96} & \textbf{28.71} & \textbf{0.758} & \textbf{224.56} & \textbf{22.20} & \textbf{0.575} & \textbf{488.89} & \textbf{23.49} & \textbf{0.625} & \textbf{522.20} \\
\bottomrule
\end{tabular}}
\vskip -0.05in
\caption{\textbf{Reconstruction comparison at high temporal compression.} We compare our method, REGEN, with MAGVIT-v2 at different compression rates on MCL-JCV and DAVIS 2019 datasets. The best results are bold-faced.
$^{\dagger}$MAGVIT-v2 (32$\times$) faces out of memory issue at 512$\times$512, due to the 3D convolution layers in decoder.
}
\label{tab:baseline}
\vskip -0.1in
\end{table*}

\begin{figure*}[t]
\begin{center}
\scalebox{1}{\includegraphics[width=\textwidth]{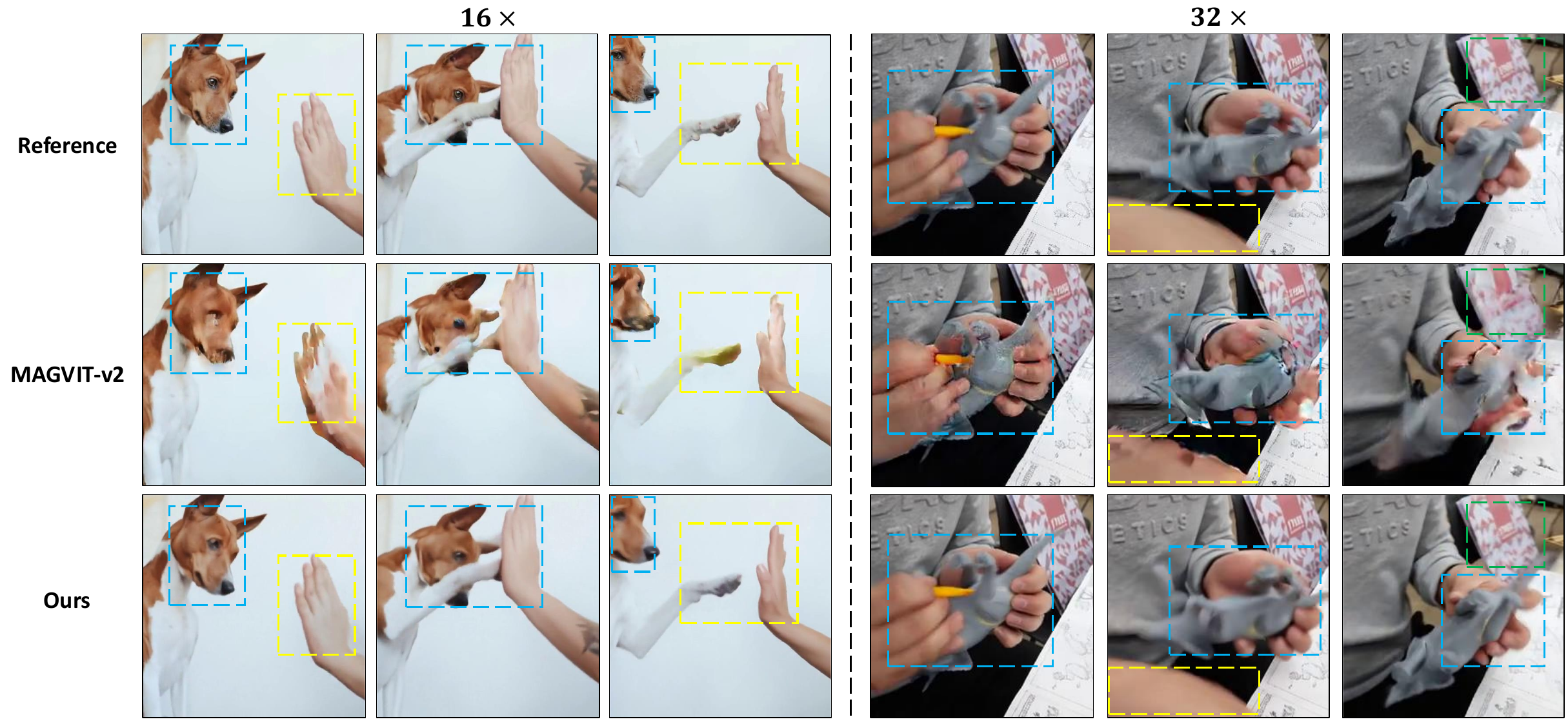}}
\end{center}
\vskip -0.25in
\caption{\textbf{Effectiveness of REGEN at high temporal compression.}
MAGVIT-v2 suffers from strong temporal artifacts in regions of high motion, such as the dog's face (left) and the toy (right). Areas enclosed in boxes show regions of maximum difference. 
}
\label{fig:recon_higher}
\vskip -0.1in
\end{figure*}

\noindent \textbf{Latent Conditioning via Content-Aware Positional Encoding.}\label{sec:siren}
The transformer model for video data usually incorporates spatiotemporal positional encoding (PE) of the space-time coordinates as additional inputs. PE is critical not only for addressing the order-invariant nature of transformers but also for capturing the spatiotemporal structure of the generated content. Existing DiT models often adopt a fixed PE scheme where the coordinate-to-embedding mapping. Such a fixed PE scheme struggles with generalizing to input sizes unseen during training \cite{yang2024cogvideox}. This is particularly problematic in our context, where we leverage DiT as the decoder since the model after training should be applied to decode the latents extracted from inputs of different resolutions and aspect ratios.

We introduce a new conditioning mechanism to address this challenge. 
The key idea is to generate the positional embedding from the conditioning latent $[z_{c}, z_{m}]$ instead of using a fixed spatial PE. Specifically, we develop a latent expansion module $C_{e}$ that takes the condition $[z_{c}, z_{m}]$ as inputs and expand it to the expanded latent $z_{e}$ same spatiotemporal dimension as the target output $x_{target}$. $z_{e}$ is then added to the token embedding and timestep embedding and input into the DiT. In this way, $C_{e}$ operates as a mini-decoder that decodes the latent into a full spatiotemporal form, and $z_{e}$ operates as a content-aware positional embedding that participates in controlling the spatiotemporal aspects in the synthesized videos.

As a positioning encoding module, $C_{e}$ is designed to map the coordinates $(x, y, t_{f})$ of a token location (after patchification) in DiT input space into an embedding vector where the mapping is conditioned on $[z_c, z_m]$. We defined such mapping as:
\begin{equation}
    C_e(x, y, t_{f} | [z_{c}, z_{m}]) = M_s(z_{c}(x,y) \ostar M_t(t_{f} | z_{m}(x,y))) 
\end{equation}
where $\ostar$ denotes the channel-wise concatenation operation. Note that the $(x, y)$ coordinates in the DiT input space match with those in the feature maps $[z_c, z_m]$ as we matched the patchification's patch size with the spatial downsampling factors. 
$M_s$ is a projector to align the channel dimension, consisting of a single linear layer and one RMSNorm \cite{zhang2019root}.
$M_t$ is a neural network sub-module that maps the time coordinate $t_f$ into a feature vector. Inspired by the success of implicit neural representation in modeling video data \cite{chen2022videoinr,mai2022motion}, we implement $M_t$ with the SIREN~\cite{sitzmann2020implicit}. We condition the mapping in $M_t$ on the motion feature $z_m$ by modulating the SIREN network with the feature values of $z_m$ at the queried $(x, y)$ location.
Thus, the positional information is integrated into the expanded latent so that we can entirely remove the original spatial and temporal positional embedding in DiT, enabling our generative decoder to be generalize across arbitrary resolutions and aspect ratios.


\section{Experiments}
\label{sec:exp}

We evaluate our encoder-generative decoder framework in three aspects: (1) effectiveness of generative decoder at various temporal compression ratios, and
(2) the compatibility of our compressed latent space for video generation.

\begin{figure*}[t]
\begin{center}
\scalebox{1}{\includegraphics[width=\textwidth]{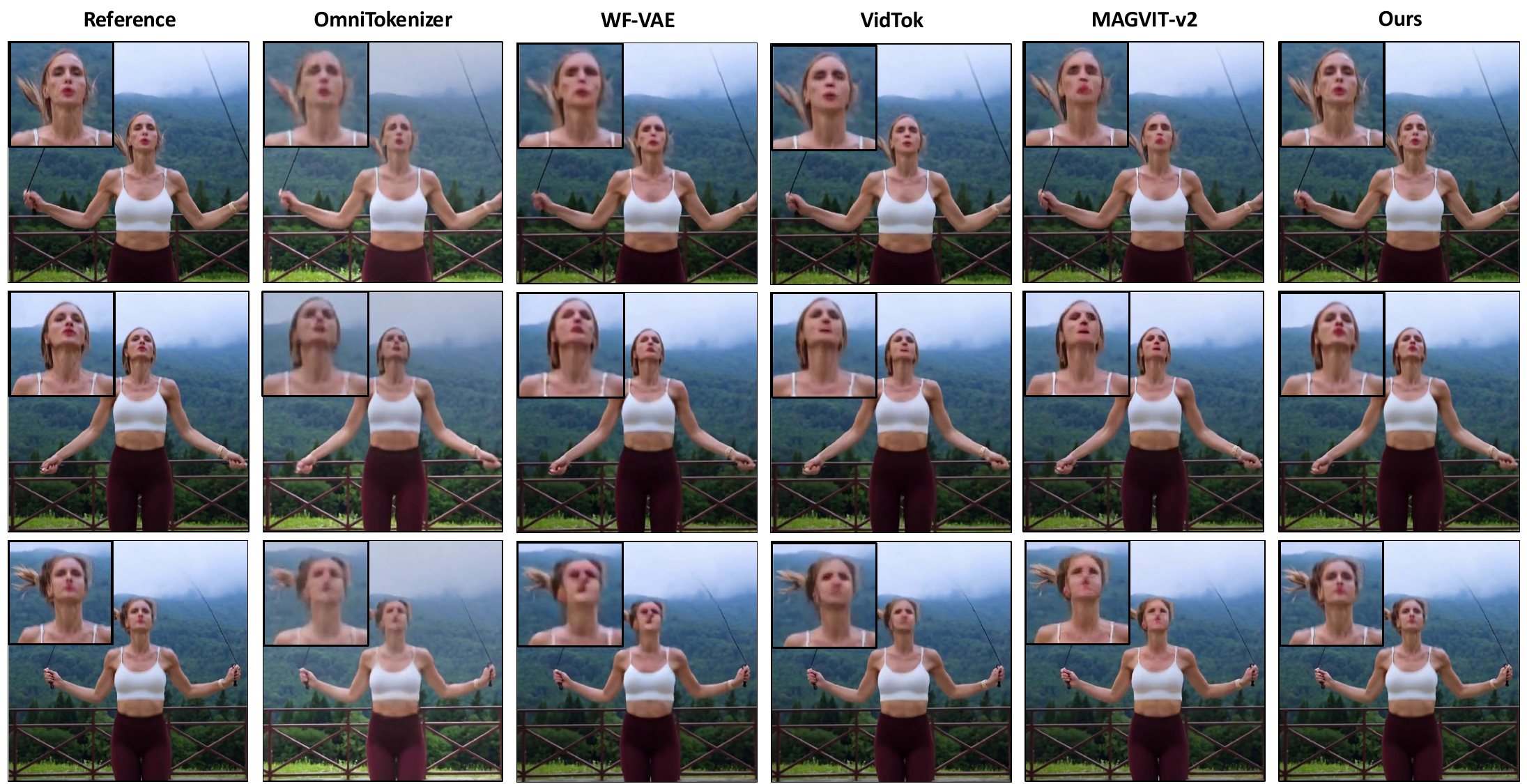}}
\end{center}
\vskip -0.23in
\caption{\textbf{Effectiveness of REGEN at base 4$\times$ temporal compression.}
Current video embedders suffer from ghosting artifacts for videos with large motion, especially in faces (last row). REGEN performs well and retains plausible spatiotemporal structures from the input.
}
\label{fig:recon}
\vskip -0.1in
\end{figure*}

\subsection{Effectiveness of REGEN as Video Embedder}

\noindent\textbf{Implementation Details.}
We configure our DiT-based decoder to match the model latency of MAGVIT-v2 \cite{yu2023language}:
it consists of 24 Transformer blocks, with 16 heads and a hidden dimension of 2048 using full satiotemporal self-attention.
We set the patch size to be 8 to accommodate the costs.
For each compression ratio, 
we train our model
for $\sim$ 100K iterations under the scenarios of reconstruction, interpolation, and extrapolation objectives. 
Our encoder shares the same architecture with our implemented MAGVIT-v2.
Please refer to the supplementary material for details.

\noindent\textbf{Evaluation.} 
Following MAGVIT-v2, we evaluate the reconstruction quality on 2 benchmarks: MCL-JCV \cite{wang2016mcl} and DAVIS 2019 (full resolution) \cite{caelles20192019}.
In both datasets, videos are rescaled and center-cropped to 256x256 and 512x512 resolution for evaluation.
We use standard quantitative metrics, PSNR, SSIM, and Fr\'{e}chet video distance (rFVD) \cite{unterthiner2018towards, ge2024content}, to examine the reconstruction quality. This covers both pixel-based and perceptual quality metrics.

\noindent\textbf{Baselines.} At 8$\times$8$\times$4 compression, we compare REGEN with SOTA 8-channel video embedders since the number of latent channels greatly affects reconstruction quality. We compare against OmniTokenizer \cite{wang2025omnitokenizer}, VidTok \cite{tang2024vidtok} and, WF-VAE \cite{li2024wf} and MAGVIT-v2 \cite{yu2023language}. 
Since the MAGVIT-v2 weights are not released, we reimplement it and train it on our dataset with over 200K iterations.
For higher compression experiments, we adapt MAGVIT-v2's design to handle different temporal compression ratios.
Note that we adjust the location of upsampling layers in MAGVIT-v2 decoder at 8$\times$8$\times$32 to avoid out of GPU memory during training and we follow the original design in other experiments to ensure the optimal reconstruction quality.

\begin{table}[t]
\centering
\scalebox{0.8}{\begin{tabular}{lcccccc}
\toprule
\multirow{2}*{Method} & \multicolumn{3}{c}{MCL-JCV} & \multicolumn{3}{c}{DAVIS 2019} \\
\cmidrule(lr){2-4}\cmidrule(lr){5-7}
& PSNR & SSIM & rFVD$\downarrow$ & PSNR & SSIM & rFVD$\downarrow$ \\
\midrule
\fineshape{} Omni & 24.63 & 0.710 & 93.35 & 23.39 & 0.628 & 152.01 \\
\ensshape{} WF-VAE  & 31.00 & 0.804 & 55.01 & 27.95 & 0.737 & 107.67 \\
\proshape{} VidTok & \underline{32.06} & \underline{0.836} & 38.85 & \underline{28.67} & \underline{0.760} & 67.24 \\
\stshape{} MAGVIT-v2 & 31.49 & 0.829 & \underline{28.63} & 28.16 & 0.758 & \underline{56.46} \\
\mixshape{} REGEN  & \textbf{32.94} & \textbf{0.857} & \textbf{22.40} & \textbf{30.25} & \textbf{0.801} & \textbf{48.38} \\
\bottomrule
\end{tabular}}
\vskip -0.05in
\caption{\textbf{Reconstruction comparison at base 4$\times$ temporal compression.} We compare REGEN with various SOTA 8-channel video embedders at 4$\times$ temporal compression on MCL-JCV and DAVIS 2019 datasets under 512$\times$512 inputs. The best results are bold-faced and the second best results are underlined.
}
\label{tab:sota}
\vskip -0.15in
\end{table}

\noindent\textbf{REGEN's effectiveness at high compression.} We investigate our primary objective, whether a diffusion-based encoder-generator framework can learn a very compact latent space with a high degree of temporal compression that can faithfully reconstruct the input much more effectively than traditional video embedders.
As shown in Tab.~\ref{tab:baseline}, MAGVIT-v2 lags behind our method on all metrics at different temporal compression rates. Moreover, the advantage of REGEN over MAGVIT-v2 widens with the increasing temporal compression ratios, which suggests that naively extending conventional video embedders to high compression rates cannot deliver satisfying results and highlights the generation ability of our decoder.
Shown in Fig.~\ref{fig:recon_higher}, MAGVIT-v2 suffers from significant degradation in reconstruction quality and often exhibits visual artifacts at high temporal compressions.
In contrast, our method maintains the reconstruction ability to a good extent. 
Moreover, our decoder is more efficient in terms of GPU memory when decoding long sequences compared to MAGVIT-v2, which is quite important in practical applications.

\begin{figure*}[t]
\begin{center}
\scalebox{0.98}{\includegraphics[width=\textwidth]{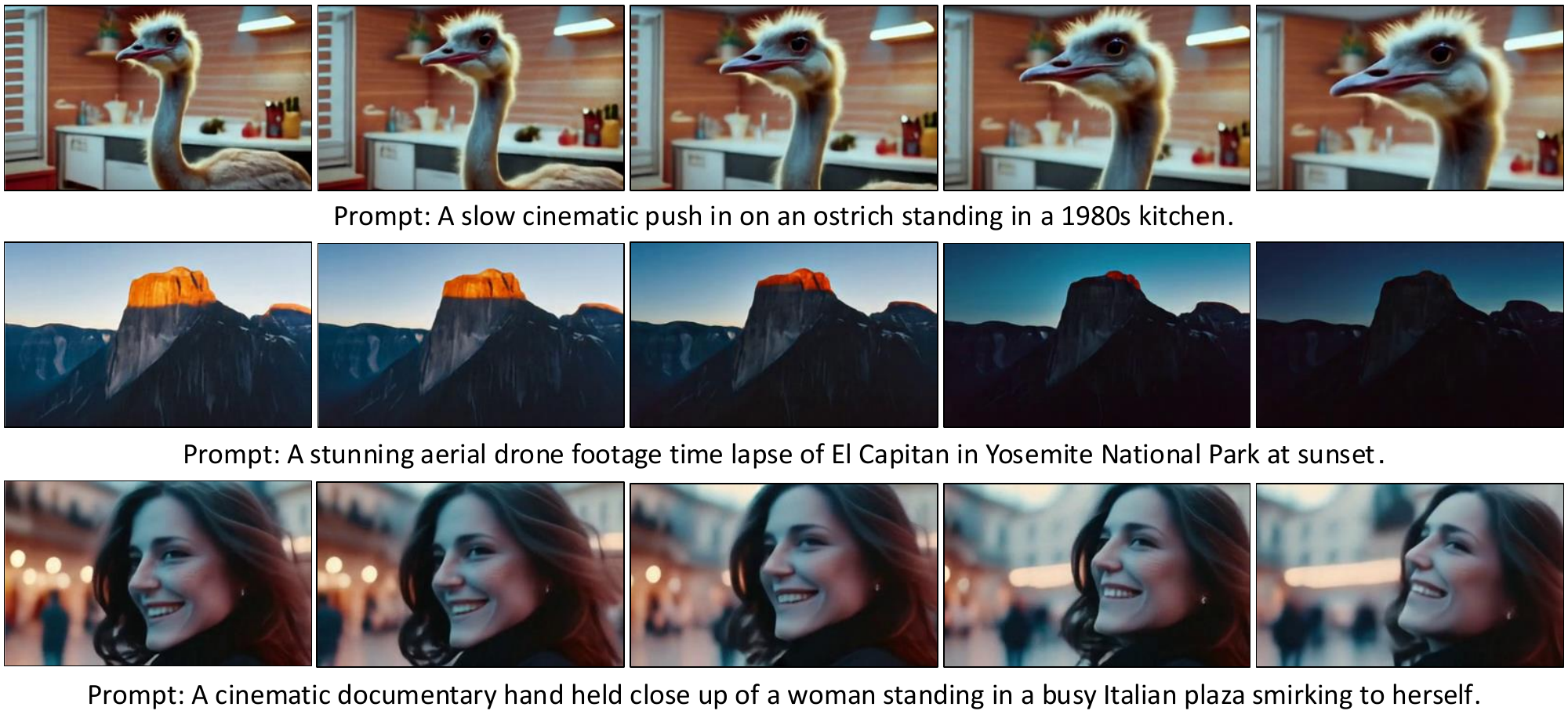}}
\end{center}
\vskip -0.25in
\caption{ 
\textbf{Text-to-video generation results on our ultra-compact latent space with 32$\times$ temporal compression.}
The latent diffusion model generates 132-frame videos with only 8 latent frames on our ultra-compact latent space, offering $\sim5\times$ reduction in the number of latent frames compared to current video embedders at 4$\times$ temporal compression.
}
\label{fig:t2v}
\end{figure*}

\begin{figure*}[t]
\vskip -0.1in
    \centering
    \begin{subfigure}[b]{0.46 \textwidth}
           \centering         \includegraphics[width=\textwidth]{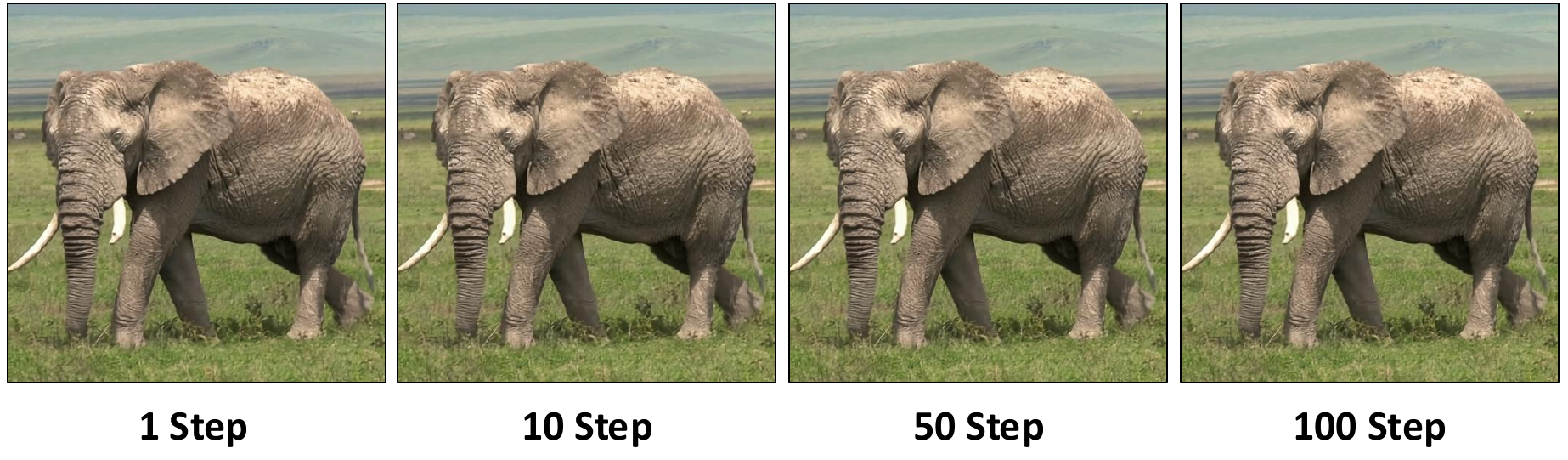}
           \vskip -0.05in
            \caption{Example reconstruction results at different sampling steps.}
            \label{fig:steps_visual}
    \end{subfigure}
    \hfill
    \begin{subfigure}[b]{0.265 \textwidth}
           \centering         \includegraphics[width=\textwidth]{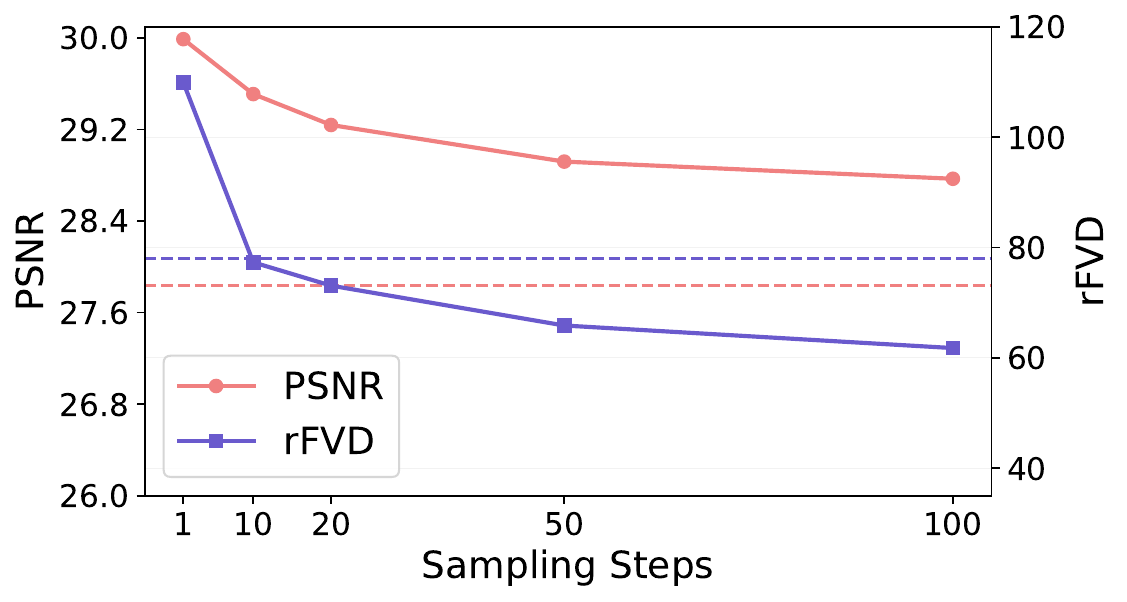}
           \vskip -0.05in
            \caption{Quantitative results on MCL-JCV.}
            \label{fig:steps_mcl}
    \end{subfigure}
    \hfill
    \begin{subfigure}[b]{0.265 \textwidth}
            \centering       \includegraphics[width=\textwidth]{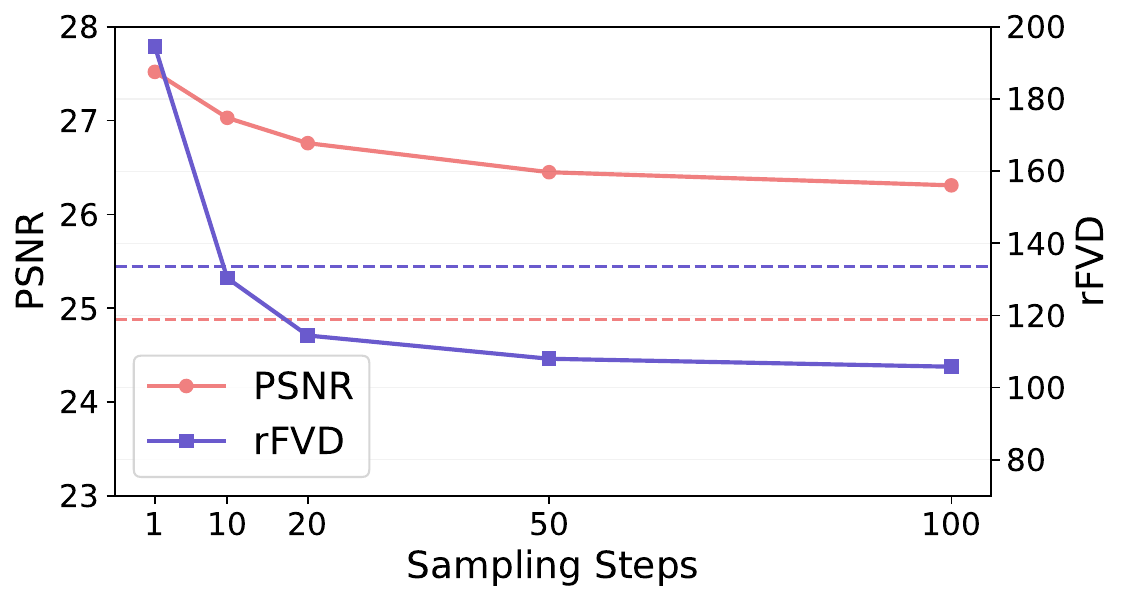}
           \vskip -0.05in
            \caption{
            Quantitative results on DAVIS 2019.}
            \label{fig:steps_davis}
    \end{subfigure}
    \vskip -0.1in
    \caption{\textbf{Reconstruction results at different sampling steps.} 
    Dash lines in (b) and (c) denote the PSNR and rFVD of MAGVIT-v2.
    Our method exhibits strong performance across varying sampling steps and
    supports few-step and even single-step sampling.}
\label{fig:step}
\vskip -0.15in
\end{figure*}

\noindent\textbf{Comparison with SOTA video embedders at base 4$\times$ temporal compression.}
Although our goal is to strive for higher temporal compression instead of SOTA performance at 8$\times$8$\times$4 compression, we compare REGEN with existing 8-channel video embedders in Tab.~\ref{tab:sota} to validate the 
effectiveness of our method. 
We observe that our implemented MAGVIT-v2 achieves compelling performance compared to recent SOTA methods VidTok \cite{tang2024vidtok}, WF-VAE \cite{li2024wf}, and OmniTokenizer \cite{wang2025omnitokenizer},
which proves the soundness of our experimental setup.
Nevertheless, all the approaches lag behind our method in all metrics, and the qualitative comparison. In Fig.~\ref{fig:recon}, we see that REGEN preserves better spatiotemporal structures from the input, especially in the human faces, which other convolution-based embedders, designed explicitly for 4$\times$ compression ratio, fail to reconstruct accurately.
We further provide a holistic comparison with other video embedders that have different latent channels in supplementary material.

\subsection{Text-to-Video Generation}

The previous experiments demonstrate that the compact video embeddings from our methods can achieve promising performance for video reconstruction. 
We now verify whether our compact latent space is friendly to text-to-video generation, which is our end objective for learning a compact latent space. 
To validate it, we train a 5B DiT-based latent diffusion model on the ultra-compact latent with 32$\times$ temporal compression for text-to-video generation. The design of the text-to-video model is based on MMDiT~\cite{esser2024scaling}. 
We first train the model with image inputs for 260K iterations, and the model is further trained with mixed image and video inputs for another 80K iterations.
We show samples of the generated videos in Fig.~\ref{fig:t2v}. 
Note that due to constraints in computing resources, we trained the model at a relatively small scale with a small number of iterations. Nevertheless, even with small-scale training, we demonstrate that the diffusion model is capable of generating plausible video content on a much more compact latent space, promising the application of our method in video generation.
Notably, the latent diffusion model generates 132-frame videos with only 8 latent frames on our ultra-compact latent space (32$\times$ temporal compression), offering $\sim5\times$ reduction in the number of latent frames compared to current video embedders that operate with 4$\times$ temporal compression, thus reducing the training and inference costs remarkably.
Please find more results in the supplementary.

\section{Discussion}
\label{sec:disc}
In this section, we address two of the main concerns with the practical usability of our transformer-based diffusion model for embedders: (1) the diffusion model inherently requires many denoising steps at inference, making it very costly, and (2) transformers are difficult to generalize to resolutions and aspect ratios unseen during training. We also provide a method to alleviate the chunking issue by leveraging the extrapolation ability of our decoder.

\noindent \textbf{Few-step and one-step sampling.}
One of the main bottlenecks with the diffusion model is that it requires a number of denoising steps, even at inference, to achieve good-quality results, which causes a lot of overhead for practical use.
However, our decoder is in fact tasked with an easier generation problem compared to text-to-video generation as it is equipped with a very strong conditioning signal provided by our latent conditioning module.
Thus, we perform inference on the same model with different numbers of steps to inspect how significantly the reconstruction quality degrades as the number of sampling steps reduces.
Fig.~\ref{fig:steps_visual} shows reconstruction examples obtained at different sampling steps using our $4 \times$ compression model. Interestingly, we observe that it is possible to reduce the number of sampling steps significantly (even down to 1-step sampling) without noticeable visual quality degradation.
We further evaluate our model at different sampling steps on all test videos from DAVIS 2019 and MCL-JCV datasets. 
Fig.~\ref{fig:steps_mcl} and Fig.~\ref{fig:steps_davis} suggest that more sampling steps lead to slightly worse PSNR and better rFVD scores, which can be attributed to the increase in sharpness in the reconstructed videos. To conclude, our diffusion decoder offers a flexible decoding scheme and can even act like a feed-forward model without utilizing external distillation, 
promising its application for practical purposes.

\begin{figure}[t]
\begin{center}
\scalebox{0.48}{\includegraphics[width=\textwidth]{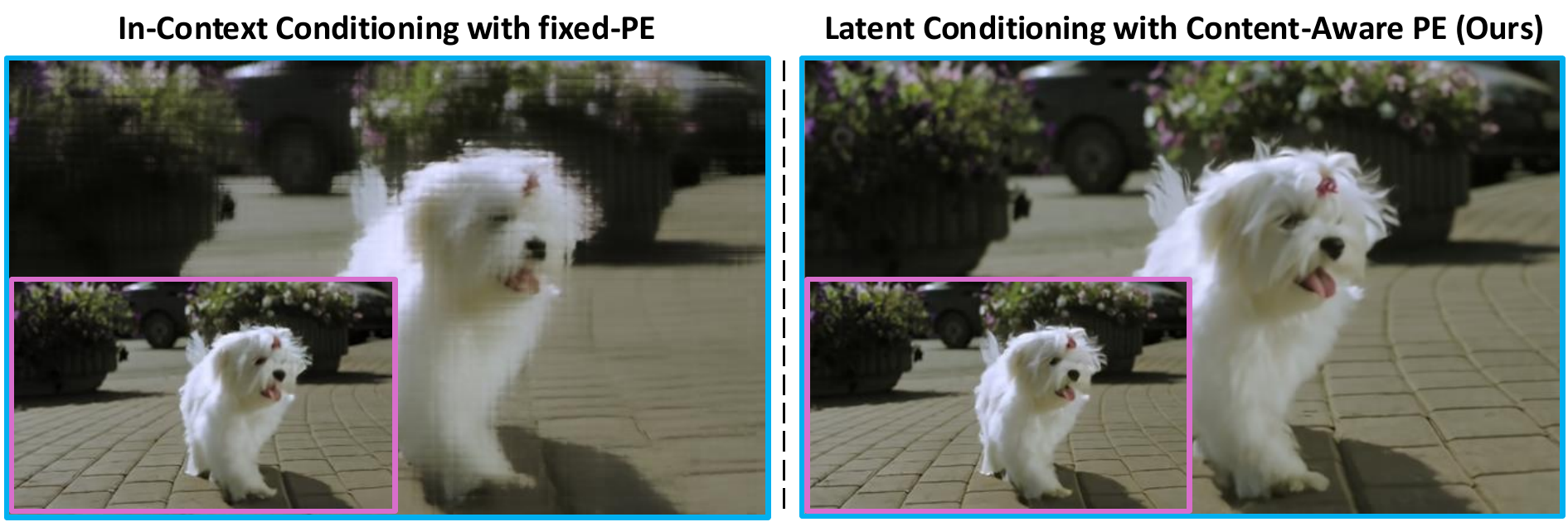}}
\end{center}
\vskip -0.22in
\caption{\textbf{Video frame reconstruction at different spatial resolutions.} 
The models are trained under 192$\times$320 input and evaluated at the resolution of 192$\times$320 (pink bounding boxes) and 384$\times$640 (blue bounding boxes).
In-context conditioning results in gridding artifacts at a larger resolution, while our method exhibits strong generalization due to the proposed content-aware PE.
}
\label{fig:res}
\vskip -0.05in
\end{figure}

\begin{table}[t]
\centering
\scalebox{0.8}{\begin{tabular}{lcccccc}
\toprule
\multirow{2}*{Method} & \multicolumn{3}{c}{\textcolor{lightcoral}{192$\times$320}} & \multicolumn{3}{c}{\textcolor{mediumpurple}{384$\times$640}} \\
\cmidrule(lr){2-4}\cmidrule(lr){5-7}
& PSNR & SSIM & rFVD$\downarrow$ & PSNR & SSIM & rFVD$\downarrow$ \\
\midrule
In-context  & 25.71 & 0.709 & 135.89 & 23.39 & 0.587 & 441.98 \\
Ours  & \textbf{26.04} & \textbf{0.720} & \textbf{128.80} & \textbf{29.41} & \textbf{0.785} & \textbf{57.01} \\
\bottomrule
\end{tabular}}
\vskip -0.05in
\caption{\textbf{Ablation of conditioning mechanism on DAVIS 2019 dataset.} In-context conditioning exhibits inferior performance compared to ours at the \textcolor{lightcoral}{training resolution} and cannot generalize well to \textcolor{mediumpurple}{unseen resolutions}. The best results are bold-faced.
}
\label{tab:ablation}
\vskip -0.1in
\end{table}

\noindent\textbf{Ablation of conditioning mechanism.}
A significant challenge with using a transformer as a decoder is that conventional transformers with fixed positional encoding (PE) can not generalize to unseen resolutions at inference. 
In contrast, our content-aware positional encoding makes REGEN inherently flexible to varying aspect ratios and resolutions at inference, unseen during training.
We verify that with an experiment where we replace our condition module with the in-context conditioning design in DiT \cite{peebles2023scalable}. This conditioning scheme retains the fixed-PE scheme like other conventional conditioning designs and performs conditioning via concatenating the conditioning tokens (\ie the encoded latents $[z_c, z_m]$ in our context) with the input tokens along the sequence dimension. 
Fig.~\ref{fig:res} shows that our method generalizes well to different resolutions.
In contrast, in-context conditioning leads to strong gridding artifacts when evaluating at higher resolutions not seen during training, severely impairing its functionality as a video decoder. 
The quantitative results in Tab.~\ref{tab:ablation} further validate this.

\begin{figure}[t]
\begin{center}
\scalebox{0.48}{\includegraphics[width=\textwidth]{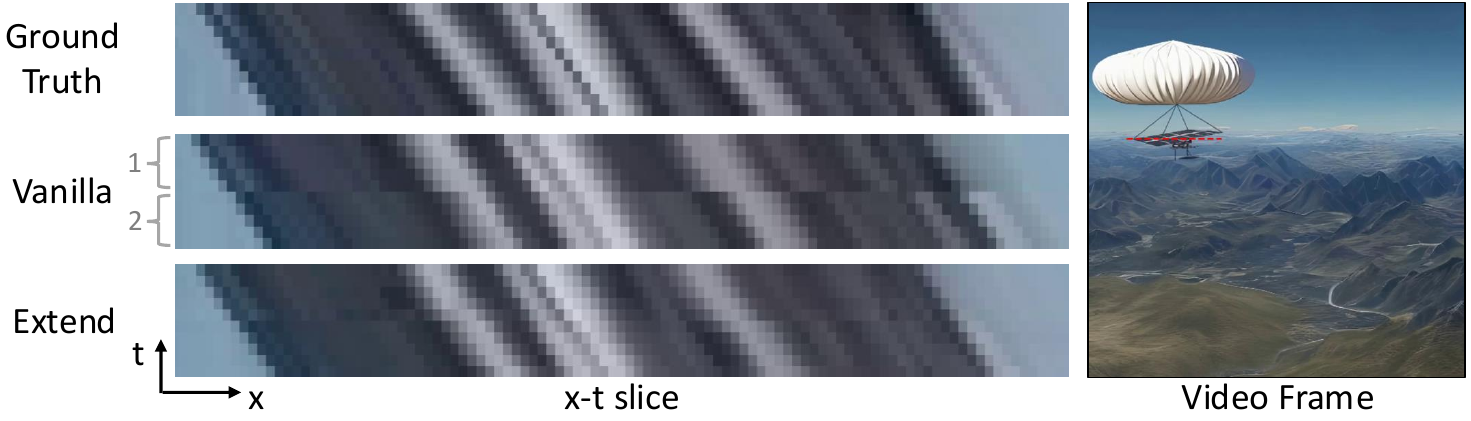}}
\end{center}
\vskip -0.25in
\caption{ 
\textbf{Alleviating the chunking issue with latent extension.}
The x-t slice is obtained by extracting a short segment (shown as the red line in the video frame) from 2 chunk frames and latent extension offers a smoother transition across the chunk boundary.
}
\label{fig:chunk}
\vskip -0.1in
\end{figure}

\noindent\textbf{Alleviating the chunking issue.}
Like other conventional video embedders, REGEN sometimes exhibits slight jumps across the intersection of two chunks due to the chunk-wise encoding scheme, which is widely adapted to process arbitrary video lengths.
To mitigate this issue, we leverage the extrapolation capability of the decoder to generate the last chunk frames based on the latent of the current chunk.
Following SDEdit~\cite{meng2021sdedit}, we utilize the predicted last frame from the previous chunk to guide the generation of the next chunk so that we can better align consecutive chunks and reduce the jumping in an autoregressive manner.
Figure~\ref{fig:chunk} demonstrates that our latent extension strategy mitigates the jumping and offers a smoother transition across the chunk boundary. Note the visible horizontal line in the middle time-slice visualization in the Vanilla decoding, which becomes much smoother when the latent extension is applied.

\section{Conclusion and Limitation}
\label{sec:concl}
In this work, we provide a new perspective on constructing the latent space for video generation. We propose that a good latent space needs to be able to generate visually plausible content that faithfully recovers an input video without needing to match the input to the pixel-perfect level of detail.
To this end, we present an encoder-generator framework, instead of the commmonly-used encoder-decoder framework trained with VAE objective, where we leverage a diffusion transformer as the decoder to ensure the latent space only captures essential semantic and structural features of the input video and the details are synthesized during decoding by the generative decoder.
We show that this approach allows us to achieve a temporal compression ratio as high as 32$\times$ with reconstruction quality far surpassing existing state-of-the-art conventional video embedders trained for such an aggressive level of compactness.

\noindent\textbf{Limitation.} 
While our method enables highly compact video embedding, there still remain some challenges and future directions:
\noindent(1) While one-step (or few-step) sampling is efficient, the training of diffusion-based decoders remains computationally expensive. Due to limited resources, we could not conduct extensive ablation on architecture configuration with the number of transformer blocks and other architectural details. Investigating this is a future direction.
(2) Although our method of latent extension could mitigate the transition challenge between chunks to great extent, it cannot fully remove the jumping issue in such a training-free manner.
(3) Our decoder operates in the pixel space, requiring a large patch size for efficiency (8 in our case), which can degrade generation quality \cite{peebles2023scalable} compared to the commonly used patch size of 1 or 2.

{
    \small
    \bibliographystyle{ieeenat_fullname}
    \bibliography{main}
}

\clearpage

\appendix

\section{Additional Video Examples}

Due to the space limit in the main paper, we provide additional video examples to validate our method and please check our \href{https://bespontaneous.github.io/REGEN/html_pages/reconstruction_4x.html}{project website} for all the results displayed in a web UI.

\section{Implementation Detail}

\subsection{Dataset} 
For video embedder training, the dataset is composed of 15 million videos and 300 million images.
For text-to-video generation, we keep the same image resources and leverage 1 million videos to construct the dataset.
The images are at the resolution of 256$\times$ with various aspect ratios, while all videos are at the resolution of 192$\times$320.

\subsection{Training}

\subsubsection{MAGVIT-v2}

We reimplement MAGVIT-v2 \cite{yu2023language} at the temporal compression ratio of $4 \times$, $8 \times$, $16 \times$, $32 \times$, and train the models at the resolution of 128$\times$128 on our dataset with over 200K iterations.
For $4 \times$ temporal compression rate, we adopt the original 4-stage design and encode 17 frames into 5 latent frames at each chunk.
For higher compression rate experiments, we add an additional stage to both encoder and decoder, and encode 17 frames into 3, 2 latent frames for temporal compression rate $8 \times$, $16 \times$, respectively.
As for the extreme $32 \times$ temporal compression rate, every 33 frames will be encoded into 2 latent frames.
We train both MAGVIT-v2 and its discriminator with the AdamW \cite{loshchilov2017decoupled} optimizer and we feed mixed images and videos to the network based on the pre-defined ratios (Image: 20$\%$, Video: 80$\%$).
All models are trained with 32 A100 GPUs and each GPU holds either 6 image inputs or 2 video sequences.

\subsubsection{REGEN}

Our method is trained at the temporal compression ratio of $4 \times$, $8 \times$, $16 \times$, $32 \times$ on our dataset with over 100K iterations.
We train our method under the setting of reconstruction, interpolation and extrapolation, and each scenario will be randomly sampled at every iteration based on pre-defined probabilities.
Specifically, we tend more towards reconstruction in the early training stage, and progressively increase the probabilities of interpolation and extrapolation with more training iterations.
Our spatiotemporal video encoder adopts the same architectural design as our implemented MAGVIT-v2 but encodes each chunk into 2 latent frames with varying compression rates.
We train the spatiotemporal video encoder and the generative decoder in an end-to-end fashion with the AdamW optimizer.
Similar with MAGVIT-v2 training, the model will be trained with mixed images and videos based on the pre-defined ratios (Image: 20$\%$, Video: 80$\%$).
All models are trained using 32 A100 GPUs, but we leverage larger batch sizes compared to MAGVIT-v2 as we did not utilize 3D convolution layers in the decoder and the GPU can process more samples.
The image batch size is set to be 28 for all experiments and the video batch size is 6, 4, 3, 1 for temporal compression ratio of $4 \times$, $8 \times$, $16 \times$, $32 \times$, respectively.

\subsection{Architecture Details}

\begin{figure*}[h]
\begin{center}
\scalebox{0.95}{\includegraphics[width=\textwidth]{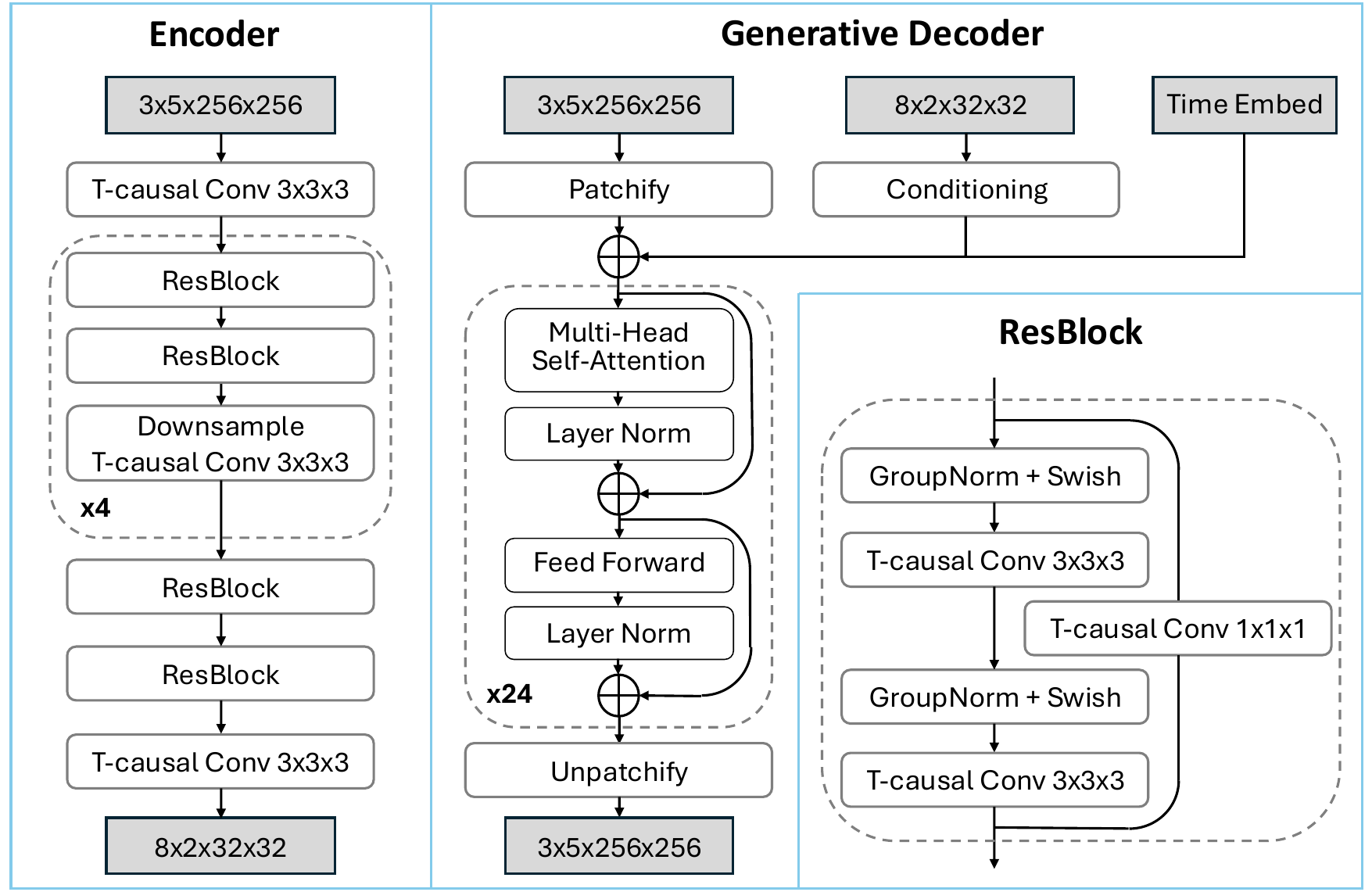}}
\end{center}
\vskip -0.15in
\caption{Architecture details of our method at $4 \times$ temporal compression rate. T-causal Conv stands for temporal causal convolution where stride = $1\times1\times1$. Downsample T-causal Conv stands for temporal causal convolution where stride will be selected from $\left\{ 1\times1\times1, 1\times2\times2, 2\times1\times1, 2\times2\times2 \right\}$, depending on the target compression ratios.
The spatiotemporal video encoder is composed of 4 stages and the generative decoder is made up of 24 diffusion transformer blocks.}
\label{fig:arch}
\vskip -0.1in
\end{figure*}

We illustrate the architecture detail of our method at $4 \times$ temporal compression rate in Fig.\ref{fig:arch}.
The spatiotemporal video encoder consists of four stages and  
it will encode every 5 raw frames into 2 latent frames with the spatial compression rate of $8 \times$.
The base channels are 128 and the channel multiplier for different stages is 1, 2, 4, 6, respectively.
The generative decoder comprises 24 diffusion transformer blocks and takes the output of encoder as the conditioning signal.
For higher compression rate experiments, we add an additional stage to the encoder while keeping the same number of diffusion transformer blocks in the decoder.
The channel multiplier is adjusted to 1, 2, 4, 6, 6, accordingly.

\subsection{Alleviating Chunking Issue with Latent Extension}

\begin{figure*}[h]
\begin{center}
\scalebox{0.95}{\includegraphics[width=\textwidth]{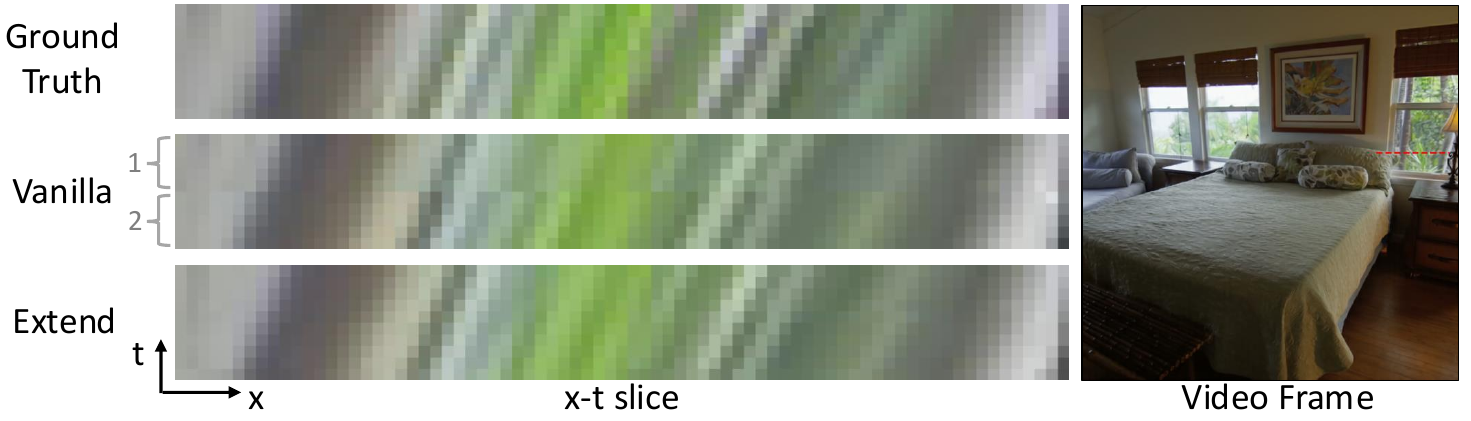}}
\end{center}
\vskip -0.25in
\caption{ 
Alleviating the chunking issue with latent extension.
The x-t slice is obtained by extracting a short segment (shown as the red line in the video frame) from 2 chunk frames. 
}
\label{fig:chunk2}
\vskip -0.2in
\end{figure*}

Following the idea of SDEdit~\cite{meng2021sdedit}, we utilize the last frame prediction from the previous chunk to guide the generation of the next chunk in an auto-regressive manner.
Specifically, given a set of latent frames, we decode the first chunk in the reconstruction setting and preserve the intermediate results of the last frame at each sampling step.
When decoding later chunks, we extend the latent by extrapolation and control the decoder to generate $T_{c}+1$ frames with the time shift of -1, where $T_{c}$ represents the length of each chunk.
In this manner, we can ensure there is one overlapped frame between consecutive chunks.
During the generation of later chunks, we leverage the intermediate results of the last frame from the previous chunk, which is also the first frame of the current chunk, to guide the generation of the current chunk so that we could mitigate the jumping issue in an auto-regressive way.
We provide an additional example in Fig.~\ref{fig:chunk2} to demonstrate the effect.

\section{Additional Comparisons with SOTA Video Embedders}

\begin{figure*}[h]
\vskip -0.1in
    \centering
    \begin{subfigure}[b]{0.48 \textwidth}
           \centering         \includegraphics[width=\textwidth]{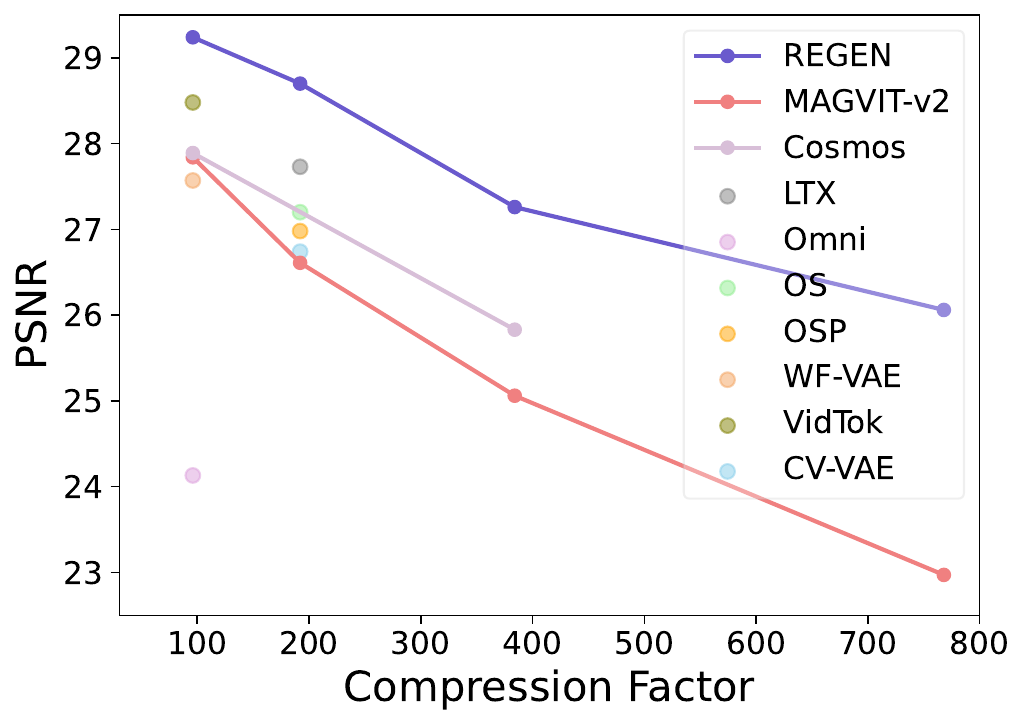}
           \vskip -0.05in
            \caption{Comparisons with state-of-the-art video embedders on MCL-JCV dataset under 256$\times$256 inputs.}
            \label{fig:mcl}
    \end{subfigure}
    \hfill
    \begin{subfigure}[b]{0.48 \textwidth}
           \centering         \includegraphics[width=\textwidth]{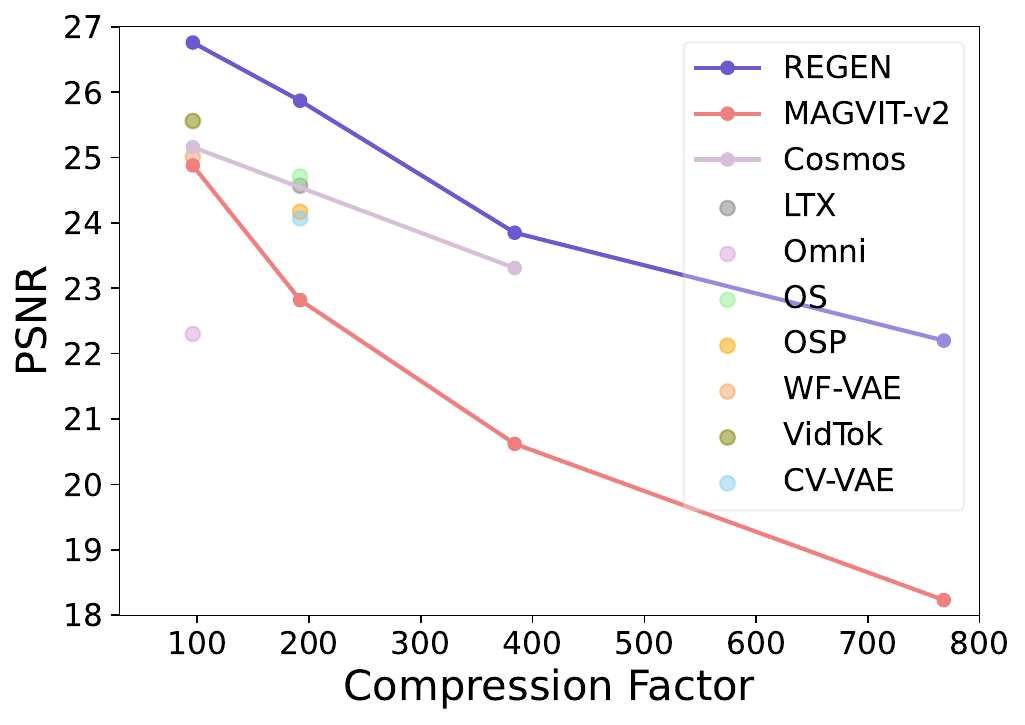}
           \vskip -0.05in
            \caption{Comparisons with state-of-the-art video embedders on DAVIS 2019 dataset under 256$\times$256 inputs.}
            \label{fig:davis}
    \end{subfigure}
    \vskip -0.1in
    \caption{Holistic comparison with state-of-the-art video embedders with different latent channel dimensions. The compression factors is calculated by $r = \frac{C\times H\times W\times (T-1)}{c\times h\times w\times t}$. REGEN obtains better PSNR compared to existing video embedders at various compression factors on different datasets.}
\label{fig:sota}
\vskip -0.1in
\end{figure*}

As many existing video embedders have different latent channel dimensions. Some video embedders, like Open-SORA (OS) \cite{opensora} and Open-SORA-Plan (OSP) only have 4 latent channels. For fairness of comparison, we compare against these SOTA embedders by calculating the compression factor following LTX-Video \cite{hacohen2024ltx}: $r = \frac{C\times H\times W\times (T-1)}{c\times h\times w\times t}$.
Shown in Fig.~\ref{fig:sota}, we further compare REGEN with Cosmos-Tokenizer \cite{agarwal2025cosmos}, LTX-Video \cite{hacohen2024ltx}, OmniTokenizer \cite{wang2025omnitokenizer}, Open-SORA (OS) \cite{opensora}, Open-SORA-Plan (OSP) \cite{pku_yuan_lab_and_tuzhan_ai_etc_2024_10948109}, WF-VAE \cite{li2024wf}, VidTok \cite{tang2024vidtok} and CV-VAE \cite{zhao2025cv}.
One can observe that our method exhibits better performance at various compression factors on both datasets, demonstrating the effectiveness of our generative decoder and the soundness of our experimental setup.

\section{Efficiency Analysis of REGEN}

For the training efficiency of latent generative models, we have the same efficiency with MAGVIT-v2 \cite{yu2023language} as we leverage the same encoder with it and decoders will not affect the training of latent generative models.

Although operating on the noise input, REGEN is equipped with a large patch size to accommodate the costs and ensure efficient inference.
Since REGEN supports one-step sampling, we measure the running time (milliseconds) of REGEN on one A100 GPU with one-step decoding and compare with MAGVIT-v2 at various compression rates under 256$\times$256 inputs.
One can observe from Tab.~\ref{tab:latency} that REGEN costs similar or fewer running times except at the compression rate of 8$\times$8$\times$32.
The reason is that we have to adjust the location of upsampling layers in MAGVIT-v2 decoder at 8$\times$8$\times$32 and shift all upsampling layers towards the end blocks to avoid out of GPU memory during training due to the causal 3D convolutional layers, even on 80GB A100s. 
Note that we avoid doing this for other compression ratios of MAGVIT-v2 as it is not an optimal design and lowers reconstruction quality, as also mentioned in MAGVIT-v2 work~\cite{yu2023language}. 
For 32$\times$ temporal compression variant, this design choice is unavoidable to ensure the decoding process fits within the available GPU memory. 
Consequently, the computational complexity of this model is lower than that of the ideal design.
Even so, this version of 32$\times$ MAGVIT-v2 cannot decode videos of even 512x512 resolution which highlights the scalability of our transformer-based decoder.

\begin{table}[h]
\centering
\scalebox{0.85}{\begin{tabular}{lcccc}
\toprule
\multirow{2}*{Method} & \multicolumn{4}{c}{Latency (ms)} \\
\cmidrule(lr){2-5}
& 8$\times$8$\times$4 & 8$\times$8$\times$8 & 8$\times$8$\times$16 & 8$\times$8$\times$32 \\
\midrule
MAGVIT-v2 & 88 & 317 & 343 & 153* \\
REGEN & 89 & 159 & 295 & 548 \\
\bottomrule
\end{tabular}}
\caption{Comparison of decoder latency at various compression rates on one A100 GPU under 256$\times$256 inputs. *MAGVIT-v2 32$\times$ has small latency because we have a different decoder design where we move all upsampling blocks toward the end layers. This is not optimal from the perspective of reconstruction quality for the decoder, but this design is unavoidable otherwise the model gives Out Of GPU Memory on 80GB A100 GPUs.
}
\label{tab:latency}
\end{table}

\section{Effectiveness of Scaling Convolution-Based Video Embedders}

While we have shown that REGEN exhibits significant advantage over MAGVIT-v2 at high temporal compression rates, e.g., 16$\times$ and 32$\times$, we further scale up the MAGVIT-v2 decoder to explore whether larger model size could help to mitigate the performance gap.
Specifically, we scale up the MAGVIT-v2 decoder through the width dimension until it matches the parameter of REGEN or reaches the GPU memory.
Tab.~\ref{tab:scale} shows that simply scaling up the model size do not always translate to better results at high compression rates, e.g., the expanded MAGVIT-v2 has worse SSIM and rFVD compared to the original version at 16$\times$ compression.
Although scaling up model size has shown some improvement at 32$\times$ compression, it still lags behind REGEN obviously, suggesting that the idea of using generative decoders to escape the compression-reconstruction trade-off is effective.

\begin{table}[h]
\centering
\scalebox{0.8}{\begin{tabular}{lcccccc}
\toprule
\multirow{2}*{Method} & \multicolumn{3}{c}{8$\times$8$\times$16} & \multicolumn{3}{c}{8$\times$8$\times$32} \\
\cmidrule(lr){2-4}\cmidrule(lr){5-7}
& PSNR & SSIM & rFVD$\downarrow$ & PSNR & SSIM & rFVD$\downarrow$ \\
\midrule
MAGVIT-v2  & 20.62 & 0.527 & 441.24 & 18.23 & 0.419 & 1080.15 \\
MAGVIT-v2$\lozenge$  & 20.83 & 0.508 & 486.48 & 18.98 & 0.437 & 1020.80 \\
REGEN  & \textbf{23.85} & \textbf{0.635} & \textbf{328.83} & \textbf{22.20} & \textbf{0.575} & \textbf{488.89} \\
\bottomrule
\end{tabular}}
\vskip -0.05in
\caption{Comparisons of expanded MAGVIT-v2 with REGEN on DAVIS 2019 dataset under 256$\times$256 inputs. We expand the MAGVIT-v2 decoder by scaling up the width dimension and $\lozenge$ denotes the expanded version. The best results are bold-faced.
}
\label{tab:scale}
\end{table}

\section{Latent Interpolation and Extrapolation}\label{sec:inter_extra}

\begin{figure*}[h]
\begin{center}
\scalebox{0.88}{\includegraphics[width=\textwidth]{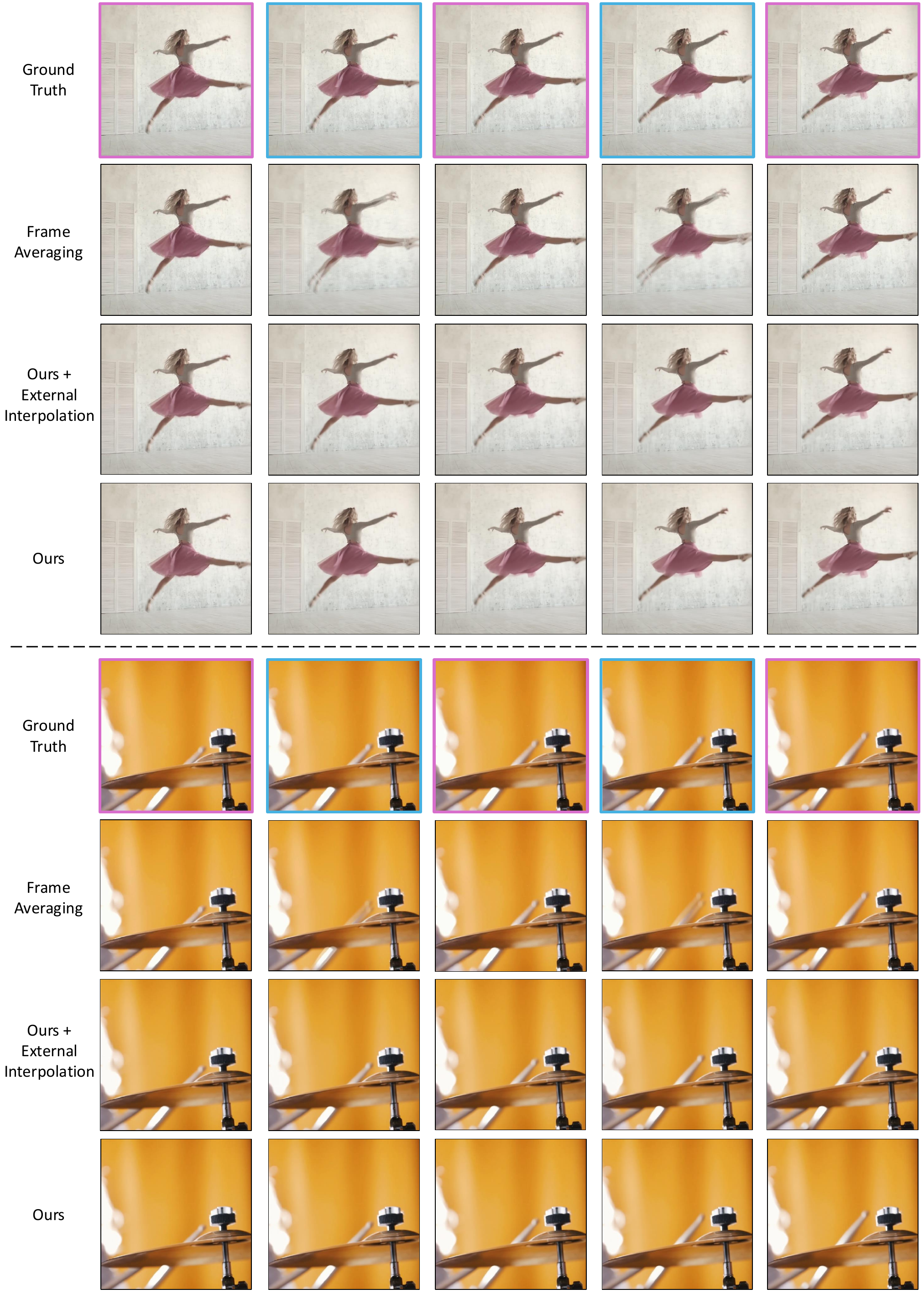}}
\end{center}
\vskip -0.15in
\caption{
2$\times$ interpolation results. Given input frames with purple bounding boxes, the model is asked to conduct interpolation to predict the frame with blue bounding box and we compare our method with frame averaging and external interpolation model.
}
\label{fig:inter}
\vskip -0.1in
\end{figure*}

\begin{figure*}[h]
\begin{center}
\scalebox{1}{\includegraphics[width=\textwidth]{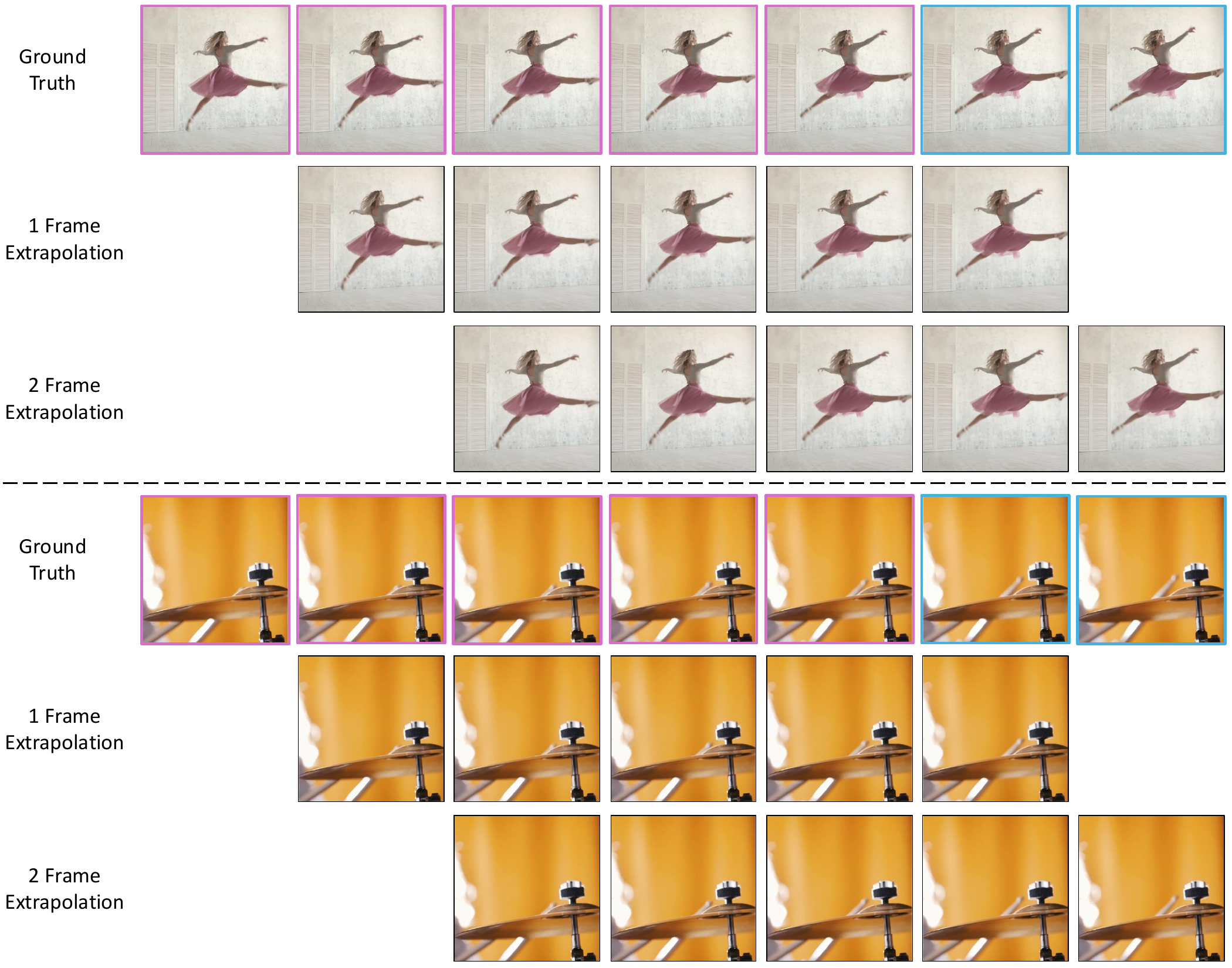}}
\end{center}
\vskip -0.15in
\caption{Forward latent extrapolation results. Given input frames with purple bounding boxes, the model is asked to conduct extrapolation to predict the future frame with blue bounding box. 
}
\label{fig:extra_forward}
\vskip -0.1in
\end{figure*}

\begin{figure*}[h]
\begin{center}
\scalebox{1}{\includegraphics[width=\textwidth]{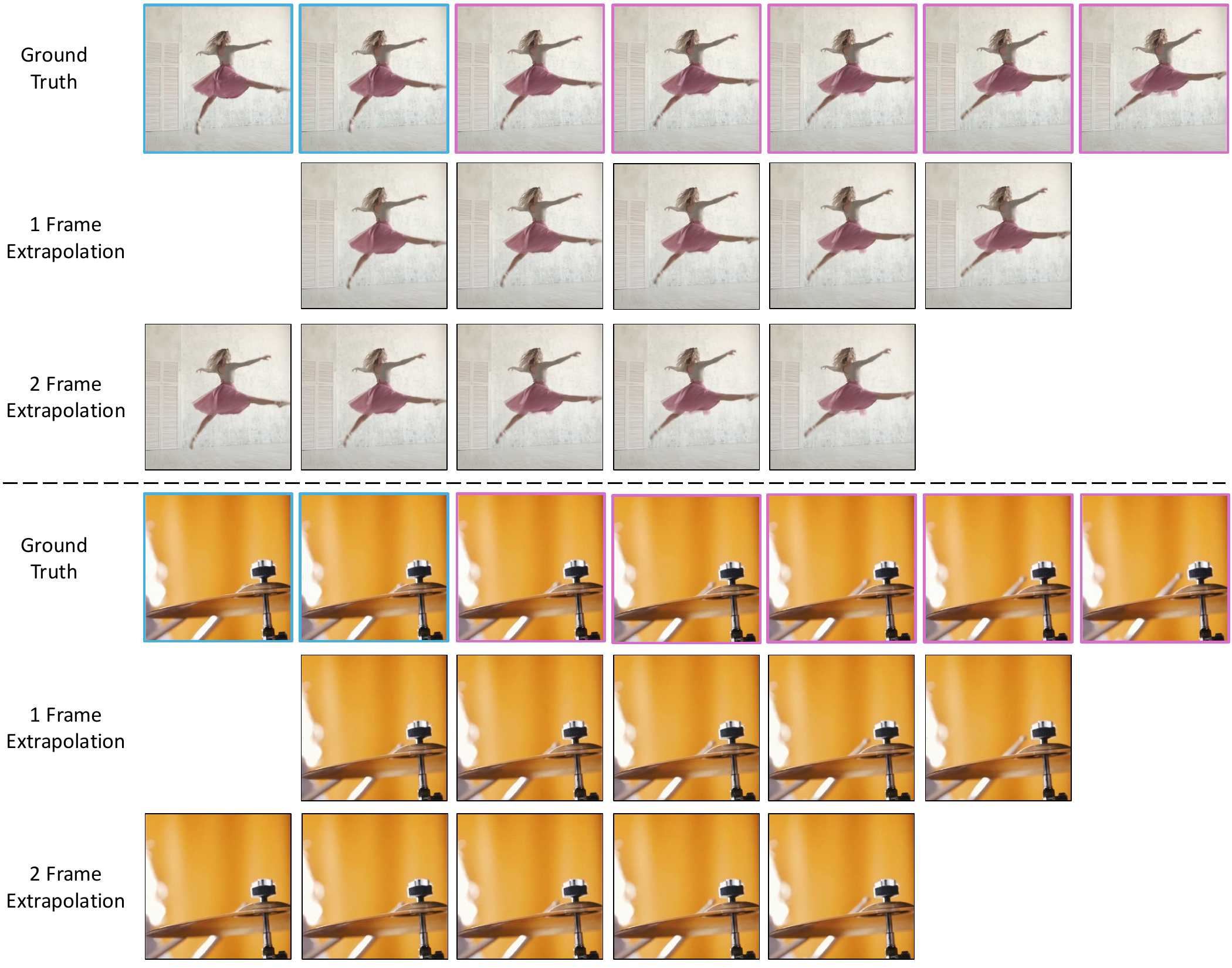}}
\end{center}
\vskip -0.15in
\caption{Backward latent extrapolation results. Given input frames with purple bounding boxes, the model is asked to conduct extrapolation to predict the past frame with blue bounding box. 
}
\label{fig:extra_backward}
\vskip -0.1in
\end{figure*}

As described in the Method, our INR-based latent conditioning module not only supports reconstruction, but can generalize to interpolation and extrapolation as well with a uniform design.
To examine the interpolation ability of our method, we compare it with two baselines:
(1) Frame Averaging: it averages the ground truth to get the interpolated frames;
(2) Ours + External
Interpolation: it applies off-the-shelf interpolation model \cite{zhang2023extracting} on our reconstructed frames.
Fig.~\ref{fig:inter} shows that simply averaging the frames results in clear artifacts on the interpolated frames, while the results of our model and EMA display a smoother transition and align well with the ground truth.
Apart from interpolation, our design supports extrapolation as well, where the model is going to predict the previous or future frames based on the given input.
Shown in Fig.~\ref{fig:extra_forward}, our approach forecasts future motion based on prior frame sequences and the results demonstrate strong alignment with the ground truth, which highlights the generation ability of our decoder.
Note that our method is able to predict the past frames as well which is shown in Fig.~\ref{fig:extra_backward}, promising the application of our method for chunk-free generation to mitigate the jumping issue.

\end{document}